\documentclass[12pt]{article}
\usepackage{graphicx}\usepackage{amsfonts}
\usepackage{amssymb}
\usepackage{pstricks}
\usepackage{pst-node}
\usepackage{pstricks-add}
\usepackage{pst-plot}
\usepackage[all,poly,2cell,ps]{xy}
\usepackage{dsfont}
\usepackage{amssymb,amsmath,latexsym}
\usepackage{natbib}
\RequirePackage[colorlinks,citecolor=red,urlcolor=blue]{hyperref}
\RequirePackage[OT1]{fontenc}

\usepackage{amssymb}
\usepackage{mathtools}

\usepackage{pstricks}
\usepackage{pst-node}
\usepackage{pstricks-add}
\usepackage{pst-plot}
\RequirePackage{hypernat}

\def\<{\langle}
\def\>{\rangle}

\def\t{\theta}
\def\<{\langle}
\def\>{\rangle}

\def\be{\begin{equation}}
 \def\ee{\end{equation}}
\def\beq{\begin{eqnarray*}}
 \def\eeq{\end{eqnarray*}}
 \def\tl{\triangleleft}

\newtheorem{theorem}{Theorem}[section]
\newtheorem{definition}{Definition}[section]

\newtheorem{lemma}{Lemma}[section]
\newtheorem{proposition}{Proposition}[section]

\newtheorem{remark}{Remark}[section]

\numberwithin{equation}{section}

\def\<{\langle}
\def\>{\rangle}

\def\<{\langle}
\def\>{\rangle}

\newcommand{\I}{{\cal I}}
\newcommand{\N}{{\cal N}}
\def\O{\mathcal{O}}
\newcommand{\M}{{\cal M}}
\newcommand{\B}{{\cal B}}

\begin{document}

\title{A local approach to estimation in discrete loglinear models}
\author{\textsc{By H\'{e}l\`{e}ne Massam\footnote{H. Massam gratefully acknowledges support from NSERC Discovery Grant No A8947.} and Nanwei Wang}\\{\it Department of Mathematics and Statistics, York University,}\\ {\it Toronto, ON M3J 1P3, Canada}}

\maketitle
\begin{abstract}
We consider  two  connected aspects of maximum likelihood estimation  of the parameter for  high-dimensional discrete graphical models:  
the existence of the maximum likelihood estimate (mle) and its computation.  

When the data is sparse, there are many zeros in the contingency table and  the maximum likelihood estimate of the parameter may not exist. Fienberg and Rinaldo (2012) have shown that the mle does not exists iff the data vector belongs to a face of the so-called marginal cone spanned by the rows of the design matrix  of the  model. Identifying these faces in high-dimension is challenging. In this paper, we take a local approach : we show that one such face, albeit possibly not the smallest one, can be identified by looking at a collection of marginal graphical models generated by induced subgraphs $G_i,i=1,\ldots,k$ of $G$. This is our first contribution.
% These models cover the given model  in the sense that the union of the interactions of the model Markov with respect to $G_i$ is equal to the interactions of the given model Markov with respect to $G$.

Our second contribution concerns the composite maximum likelihood estimate. When the dimension of the problem is large, estimating the parameters of a given graphical model through maximum likelihood is challenging, if not impossible. The traditional approach to this problem has  been local with the use of composite likelihood based on local conditional likelihoods.
 A more recent development is to have the components of the composite likelihood be marginal likelihoods centred around each $v$. We first show that the estimates obtained by consensus through local conditional and marginal likelihoods are identical. We then study the asymptotic properties of the composite maximum likelihood estimate when both the dimension of the model and the sample size $N$ go to infinity.
 
% Existence and computation of the mle are clearly connected also at the local level. Indeed if one or more of the local maximum likelihood estimates do not exist,  standard optimization software  may give, without warning, erroneous estimate of the non existent local mle  and this, of course, results in inaccurate estimates of the composite mle.

\end{abstract}

%62H17 (Primary),  
%62M40.

\vspace{4mm}


\section{Introduction}
Let $V=\{1,\ldots,p\}$ be a finite index set. We consider $N$ individuals that we classify according to criteria or variables $X_v, v\in V$. We assume that for each $v\in V$, $X_v$ take its values in a finite set $I_v.$ 
Let $G=(V,E)$ be an undirected graph where  $E$ is the set of undirected edges $(i,j)\in V\times V$.
We  assume that independences and conditional independences between the variables $X_v, v\in V$  are represented by $G$ in the following way:
$$X_{v_1}\perp X_{v_2}\;|\; X_{V\setminus \{v_1,v_2\}}\;\;\mbox{if}\;\;(v_1,v_2)\not \in E.$$
Thus the distribution of $X=(X_v, v\in V)$ belongs to a discrete graphical model Markov with respect to $G$. 
The data are gathered in a $p$-dimensional contingency table with cells $I=\prod_{v\in V}I_v$. The parameters of this model are the cell probabilities or, equivalently, the loglinear parameters $\theta$ in \eqref{basic} below if we write the density of the cell counts under its natural exponential family form
 \begin{eqnarray}
\label{basic}
f(t; \theta)=\exp \{\langle \theta, t \rangle-Nk(\theta)\}\;.
\end{eqnarray}
Here $t$ is a vector of marginal cell counts and $\langle \theta, t \rangle$ denotes the inner product of $t=t(x)$ and the canonical loglinear parameter $\theta$. 
Discrete loglinear  models are widely used in many scientific areas and  two aspects of these models have been the topic of much research. The first is the existence of the maximum likelihood estimate (henceforth abbreviated mle) of the parameters. And the second is that of the computation or approximate computation of the mle. These two aspects are connected as we shall see below. Our contribution in this paper is two-fold and concerns these two aspects.

 The mle of the parameter is said to exist if all  cell probability estimates are strictly positive.  The reader is referred to Fienberg \& Rinaldo (2007) and Fienberg \& Rinaldo (2012) for a  complete list of references on the topic and a  most interesting historical account of the developments. It has been known for a long time (see Birch, 1963 and Haberman, 1974) that the nonexistence of the mle is due to the presence of zeros in the contingency table. Zeros can exist even when $p$ is small, but they  occur particularly often  when $p$ is large and the sample size $N$ is relatively small. Whether  the mle exists or not is determined by the position of the data vector $t$ in the convex hull  of the support of the measure $\mu$ generating the hierarchical loglinear model \eqref{basic}. Eriksson et al. (2006) were the first to express a necessary and sufficient condition for the existence of the mle in these terms. They showed that the mle exists if and only if the data vector belongs to the interior of what they call the marginal cone, i.e., the cone $C$ generated by the convex hull of the support of $\mu$. Fienberg and Rinaldo (2012) showed that this necessary and sufficient condition is valid for all sampling schemes (Poisson, multinomial or product multinomial). The marginal cone  $C$ is a polyhedral cone and, thus, in order to determine whether  the mle exists, one has to identify whether the data vector belongs to one of its faces.
In the supplementary material of their paper, Fienberg and Rinaldo (2012)   gave several linear programming algorithms to identify the smallest face of the cone containing the data vector. Practically, however,   these algorithms cannot be implemented in high dimensions.   

Our first contribution in this paper is to provide, for high-dimensional problems, a way to identify a face of $C$ containing the data vector $t$.
To do so, we take a local approach. We consider a finite  collection of   subgraphs $G_{A_i}$  of $G$ induced by subsets $A_i, \;i=1,\ldots, k$ of $V$. We solve each local problem, that is, we find the smallest face $F_{A_i}$ containing the vector of marginal cell counts $t_{A_i}$ in the marginal model Markov with respect to $G_{A_i}$. This face can be extended to a face $F_i$ of $C$ containing $t$. In Theorem \ref{main1}, we show that $\cap_{i=1}^k F_i $ is a face of the marginal cone $C$ of the global model
containing the data vector $t$.
 The face $\cap_{i=1}^k F_i $ is not necessarily  the smallest one containing $t$, but  knowing that $t$ belongs to $\cap_{i=1}^k F_i $ tells us that the mle does not exist and gives us a model of  dimension smaller than the dimension of ${\cal M}$ for which we can attempt to evaluate the mle. Of course, this is true only if  $\cap_{i=1}^k F_i$ is not equal to the relative interior of $C$. If it is, our procedure yields no information. In our simulations, if the data belonged to a face of $C$, we never had this situation. In fact, we found that the face  containing $t$ was always equal to the intersection $\cap_{i=1}^k F_i $. This is probably due to the fact that  a simulated data vector will  rarely fall on  a face of small dimension which would not show up as  $\cap_{i=1}^k F_i $. 
 %It is, however, straightforward to build a counterexample showing that a face of $C$ is not necessarily included in the intersection $\cap_{i=1}^k F_i $ whichever way $G$ is split (see section 3.3). \vspace{3mm}

 Our second contribution in this paper concerns the  composite maximum likelihood estimates of $\theta$ in \eqref{basic}.  The mle of $\theta$ is that value of the parameter $\theta$ that maximizes the likelihood  or, equivalently, the loglikelihood function $l(\theta)=\langle \theta, t \rangle-Nk(\theta)$.
The log partition function or cumulant generating function $k(\theta)$  is usually intractable when the dimension of the model is large. 
  As a consequence, even though the likelihood function is a convex function of $\theta$, the traditional convex optimization methods using the derivative of the likelihood cannot be used. Approximate techniques such as variational methods (see Jordan et al., 1999, Wainwright and Jordan, 2008) or MCMC techniques (see Geyer, 1991, Snijders 2002) have been developed in recent years. More recently still, work has been done on a third type of approximate techniques based on the maximization of    composite likelihoods (Lindsay, 1988). For a given graphical model and a given data set $x^{(1)},\ldots, x^{(N)}$, a composite likelihood is  typically the product of local conditional  likelihoods, coming from the local conditional probability of $X_v$ given $X_{\N_v}$, which we can write  as
  \begin{equation}
  \label{cond}
  L^{PS}(\theta)=\prod_{v\in V}\prod_{k=1}^Np(X_v=x_v^{(k)}|X_{\N_v}=x^{(k)}_{\N_v};\theta)
 \end{equation}   
  where $\N_v$ indices the set of neighbours of $v$ in $G$.
  For work using  this type of techniques applied to discrete graphical models, the reader may refer to  
  %Sutton and McCallum (2007), 
  %for piecewise pseudolikelihood,  
  %Asuncion, Liu, Ihler and Smyth (2010), Boyd et al. (2010), Dillon and Lebanon (2010),
  % for composite likelihood, 
 % Bradley and Guestrin (2012)
  % for asymptotics for statistical and computational efficiency  
  Liu and Ihler (2012) and references therein.
%  Pseudolikelihood  has also been used for discrete graphical model selection by Ravikumar, Wainwright and Lafferty (2010).

   In the last two years, for Gaussian graphical models, Wiesel and Hero (2012) and Meng et al. (2013, 2014) have introduced composite likelihood estimation where the composite likelihood 
%\begin{equation}
%\label{marg}
%L_c^{\M}(\theta)=\prod_{v\in V}\prod_{k=1}^Np(X_v=x_v^{(k)},X_{\N_v}=x^{(k)}_{\N_v};\theta^{v,\M_{l,v}}),\;\;l=1,2,
%\end{equation}   
is the product of local convex "relaxed" marginal (rather than conditional) likelihoods coming from  $p(X_v,X_{\N_v})$ rather than from $p(X_v\mid X_{\N_v})$.  For discrete graphical models, Mizrahi et al. (2014a) have proposed a similar marginal approach, taking a clique of $G$ and its neighbourhood. In the papers mentioned above  using  the composite likelihood, from either marginal or conditional local likelihoods, the global composite mle is  obtained by "consensus", i.e. by combining the  local mle's  obtained from the various $v$-local likelihoods. The value of the global composite mle
 is therefore a function of the values of the mle of the $v$-local likelihoods.

In this paper, we extend the marginal approach of Meng et al. (2013, 2014) to discrete graphical models  and  give two results. The first, Theorem \ref{equal} states that the composite mle obtained through this marginal approach is equal to the composite mle obtained through the  conditional approach. We conclude that there is therefore no point in using the more computationally complex marginal approach to compute the composite mle. Our second  result, Theorem \ref{mbound} gives the rate of convergence of the composite mle to the true value $\theta^*$ of the parameter when both $p$ and $N$ go to infinity. In Theorem \ref{global}, we give the rate of convergence of the global mle, under the same conditions and compare the two rates of convergence.

%These local marginal models are  "relaxed" in the sense that the local likelihood is made convex in the parameter by adding some interactions. Meng et al.(2013, 2014) introduce this relaxed marginal approach,  for graphical  Gaussian model,  in order to obtain increased accuracy of the composite mle. {\bf talk about Mizrahi here}
\vspace{2mm}

The  existence  and  computation of the mle are of course connected, 
if a  local mle does not exist,  the composite mle cannot be computed by consensus and probably does not exist. An optimization routine may not signal that the mle does not exist and will usually give a fallacious number for the local mle
 In this case, running  a convex optimization routine on each local problem without looking at the details of each maximization may  lead to an inaccurate estimate (and erroneously lead to think that the two composite mle, obtained from local marginal and conditional likelihoods, are different). 
\vspace{2mm}
 
 The remainder of this paper is organized as follows.  In the next section, we introduce our notation for the discrete graphical model. In Section 3, we prove our result on the identification of a  face containing the data vector. In Section 4,  we study the local marginal likelihood estimator and its relationship to the local conditional likelihood estimator. In Section 5, we give its asymptotic properties. The proof of the theorems are given in the Appendix.

\section{Preliminaries}
\subsection{ Discrete graphical and hierarchical loglinear models}
Let $p, V$  and  $X=(X_v,\;v\in V)$ be as described in Section 1 above. If $N$ individuals are classified according to the $p$ criteria, the resulting counts are gathered in a contingency table such that
$$I=\prod_{v\in V}I_{v}$$
is the  set of cells $i=(i_v,\;v\in V)$.  For $D\subset V$, $i_D$ denotes the marginal cell $i_D=(i_v, v\in D)$ with $i_v\in I_v$.
Let $\mathcal{D}$ be a family of non empty subsets of $V$ such that $D\in \mathcal{D}$, $D_1\subset D$ and $D_1\not = \emptyset$  implies $D_1\in \mathcal{D}.$ In order to avoid trivialities we assume $\cup_{D\in \mathcal{D}}D=V.$  The  family $\mathcal{D}$ is called   the generating class of the hierarchical loglinear model. We denote by $\Omega_{\mathcal{D}}$ the linear subspace
of  $y\in {\mathbb{R}}^I$  such that there exist functions $ \theta_D\in {\mathbb{R}}^I$ for $D\in \mathcal{D}$ depending only on $i_D$ and such that $y=\sum_{D\in \mathcal{D}}\theta_D$, that is
\begin{eqnarray}
\label{omegad}\nonumber
\Omega_{\cal D}=\{y\in {\mathbb{R}}^I:\;\exists\theta_D\in {\mathbb{R}}^I, D\in {\cal D}\;\mbox{such that}\; \theta_D(i)=\theta_D(i_D)\;\mbox{and}\; y=\sum_{D\in {\cal D}}\theta_D\}
\end{eqnarray} 
The hierarchical  model generated by ${\cal D}$ is the set of probabilities $p=(p(i))_{i\in I}$ on $I$ such that $p(i)>0$ for all $i$ and such that $\log p\in \Omega_{\cal D}.$

The class of discrete graphical models Markov with respect to an undirected graph $G$ is a subclass of the class of hierarchical discrete loglinear models. Indeed, let $G=(V, E)$ be an undirected graph where $V$ is the set of vertices and $E\subset V\times V$ denotes the set of undirected edges. We say that the distribution of $X$ is Markov with respect to $G$ if $(v_1,v_2)\not \in E$ implies
$$X_{v_1}\perp X_{v_2}|\; X_{V\setminus \{v_1,v_2\}}.$$
 Let ${\cal D}$ be the set of all cliques (not necessarily maximal) of the graph $G$.
If the distribution of $X=(X_1,\ldots,X_p)$ is Multinomial($1, p(i), i\in I$)  Markov with respect to the graph $G$, and if we assume that all $p(i), i\in I,$ are positive, then, by the Hammersley-Clifford theorem, $\log p( i)$ is a linear function of parameters dependent on the marginal cells $i_D, D\in {\cal D}$ only, and therefore the graphical model is a hierarchical loglinear model with generating set the set ${\cal D}$ of cliques of $G$. The reader is referred to Darroch \& Speed (1983), Lauritzen (1996) or Letac \& Massam (2012) for a detailed description of the hierarchical loglinear model and the subclass of discrete graphical loglinear models.

 We now set our notation and recall some basic results  for discrete hierarchical loglinear models. The following notation and results can be found in Letac \& Massam (2012) and the corresponding supplementary file.

Among all the values that $X_v$ can take in $I_v, v\in V$, we call one of them $0$. For a cell $i\in I$, we define its support $S(i)$  as 
 $$S(i)=\{v\in V\ ; \ i_{v}\neq 0\}$$
and we define also the following subset  $J$ of $I$  
 \be \label{j} J=\{ j\in I,\ \ S(j)\in \mathcal{D}\}.\ee 
 From here on, we will call this set the $J$-set of the model.
 For $i\in I$ and $j\in J$, we define the symbol
 $$j\tl i$$
  to mean that $S(j)$ is contained in $S(i)$ and that $j_{S(j)}=i_{S(j)}.$  The relation   $\tl$ has the property that if $j,j'\in J$ and $i\in I$, then 
 \begin{eqnarray*}\label{TL}j\tl j'\ \ \mbox{and}\ \ j'\tl i\Rightarrow j\tl i.\end{eqnarray*}
 The log linear parametrization that we use for the multinomial is the so-called baseline parametrization with general expression, for $i\in I, S(i)=E\subset V$, 
 \be
 \label{gentheta}
 \theta_i=\sum_{F\subset E} (-1)^{|E|-|F|}\log p(i_F,0_{V\setminus F})\;.
 \ee
 With the notation above, in Proposition 2.1 of Letac and Massam (2012), it is shown  that for $i\not \in J,\;\theta_i=0$ and that
 \begin{eqnarray}
 \label{thetaj}
 \theta_j&=&\sum_{j'\in J,\; j'\tl j}(-1)^{|S(j)|-|S(j')|}\log \frac{p(j')}{p(0)},\;j\in J\nonumber\\
\label{pi}  \log p(i)&=&\theta_0+\sum_{j\in J, j\tl i} \theta_j,\;i\in I\label{pi}\\
 \label{p0} \log p(0)&=&\theta_0.\label{p0}
 \end{eqnarray}
 One then readily derives the density of the multinomial M$(N, p(i), i\in I)$ of the cell counts $\underline{n}=(n(i), i\in I)$, Markov with respect to $G$ to be, up to a multiplicative constant, equal to
 %\be \label{base} f(n(i),i\in I)=\exp \langle \theta, \underline{n}\rangle -N k(\theta) \ee
% where 
 %$$\langle \theta, \underline{n}\rangle=\sum_{j\in J}\theta_jn(j_{S(j)}), \;\;k(\theta)=\log p(0)^{-1}=\log \Big(\sum_{i\in I}\exp \sum_{j\in J, j\tl i} \theta_j\Big).$$
 %Clearly exponential family \eqref{base} for $\theta\in R^{|J|}$ is of dimension $|J|$ and, up to a multiplicative constant,  it is the set of distributions of $t=(n(j_{S(j)}), j\in J)$ with density 
 \begin{equation}
 \label{ftheta}
 f(t;\;\theta)= \exp \{\langle t, \theta \rangle-Nk(\theta)\},\; \theta\in R^{J}
 \end{equation}
 with $\theta=(\theta_j,j\in J),\; t=t(\underline{n})=(t(j),\; j\in J)$  where $t(j)=n(j_{S(j)})$ are the $j_{S(j)}$-marginal cell counts
 %, $\langle a,b\rangle$ indicates the inner product of $a$ and $b$ in $R^{J}$   
  and 
  \begin{equation}
  \label{k}
  k(\theta)=\log \Big(\sum_{i\in I} \exp \sum_{j\in J, j\tl i} \theta_j\Big)=\log \Big(1+\sum_{i\in I\setminus\{0\}} \exp \sum_{j\in J, j\tl i} \theta_j\Big)\;.
  \end{equation}
 For $\theta\in R^{J}$, these distributions form a natural exponential family of dimension $J$ generated by a measure $\mu$ which we will now identify. Let $e_j, j\in J$ be the canonical basis of $R^{J}$ and, for $i\in I$, let 
  $$f_i=\sum_{j\in J, j\tl i}e_j.$$
 Then \eqref{pi} and \eqref{p0} can be written in  matrix form as 
  \begin{eqnarray}\label{HXtilde}
\log p&=&A\tilde{\theta}
\end{eqnarray} 
where $\tilde{\theta}^t=(\theta_0, \theta^t), $ $A$ is an $(|I|)\times(1+|J|)$ matrix. We call $A$ the design matrix of the model. The rows of $A$ are indexed by $i\in I$ and equal to  $\tilde{f_i}^t=(1,f_i^t)\in {\mathbb{R}}^{J+1}$.  It is immediate to see that the Laplace transform of the generating measure $\mu$ is
$$e^{k(\theta)}=\sum_{i\in I}e^{\langle \theta, f_i\rangle}$$ and therefore the measure $\mu$ generating \eqref{ftheta} is
\begin{equation}\label{mu}
\mu(dx)=\sum_{i\in I}\delta_{f_i}(x).
\end{equation}
  This exponential family is concentrated on the convex hull of $f_i, i\in I$, which is a bounded set of  ${\mathbb{R}}^J$, and therefore the set of parameters $\theta$ for which $L$ is finite is the whole space ${\mathbb{R}}^J.$ From the definition of $X$, $f_i, i\in I$ and $t=(t(j_{S(j)}, j\in j)$, it is easy to see that
  $(N,t(j),\;j\in J)^t=A^tn=\sum_{i\in I}n(i)\tilde{f}_i$ and the vector of sufficient statistics $t$, which we also write as $t=t_J$ to emphasize its length,  is such that
 \begin{equation}
 \label{tj}
 \frac{t_J}{N}=\Big(\frac{t(j)}{N},j\in J\Big)^t=\sum_{i\in I\setminus \{0\}}\frac{n(i)}{N}f_i=\sum_{i\in I}\frac{n(i)}{N}f_i
 \end{equation} 
 belongs to the convex hull of  $(f_i)_{i\in I}$.
 The $(f_i)'s$ are the extreme points of the closure of the convex hull of the $f_i, i\in I$.
 \vspace{2mm}
 
 Fienberg and Rinaldo (2012), Theorem 3, show that the mle of $\theta$ in \eqref{ftheta} exists and is unique if and only if the canonical statistic vector $t$  belongs to the relative interior of the cone $C$ with apex $f_0$ and generated by $f_i, i\in I\setminus \{0\}.$  Therefore, if the mle does not exist, the data vector $t$ must belong to one of the facets of $C$. Following Eriksson et al. (2006) and Fienberg and Rinaldo (2012), we will call $C$ the marginal cone of the model.
 The reader  is referred to Letac and Massam (2012) for examples of the notions given above.

\section{Faces containing the data vector $t$}
\subsection{Finding the smallest face containing $t$}
Let $C$ be a closed convex polyhedral cone. A set $F\subset C$ is said to be a face of $C$ if for all $x\in F$, any decomposition of the form $x=y+z$ with $y, z\in C$  implies $y,z\in F$.
Given $g\in R^{J}$, the inequality $\langle g, x\rangle\geq 0$ is said to be valid for $C$ if it holds for every $x\in C$. Then the set
$$F_g=\{x\in C:\; \langle g, x\rangle= 0\}$$
is called the face of $C$ governed by $g$. Every face of $C$ arises in this manner.
 There is only one face of dimension 0 and  that is $\{f_0\}$. The faces of dimension 1 are the extreme rays $\{\lambda f_i,\;\lambda>0\}$ for each $f_i, i\in I\setminus \{0\}$. Since a convex set is the convex hull of its extreme points, a face $F$ of $C$ will be defined by a set
$$f_i, i\in {\cal F}\subset I.$$
To identify the smallest face $F$ of $C$ containing the data vector $t\in R^{J}$, we will have to identify the subset ${\cal F}$ of $I$ defining that face  or equivalently we will have to identify a $g\in R^{J}$ such that
\begin{eqnarray*}
\langle g, f_i\rangle&=&0,\;\forall i\in {\cal F}\\
\langle g, f_i\rangle&>&0,\;\forall i\in \I\setminus {\cal F}
\end{eqnarray*}
Let $I_+$ be the subset of $\I$ with strictly positive cell counts $n(i)>0$. We have the following lemma which shows that ${\cal F}\supset I_+$.
\begin{lemma}
\label{i+}
The sufficient statistic $t$ belongs to the face $F_g$ of $C$ governed by $g$ if and only if $f_i\in F_g$ for all $i\in I_+.$
\end{lemma}
The proof is obvious if we write that $t\in F_g\Leftrightarrow \langle t,g\rangle=0\Leftrightarrow\sum_{i\in \I_+}\frac{n(i)}{N}\langle f_i,g\rangle=0\Leftrightarrow \langle f_i,g\rangle=0\;\forall i\in I_+.$
The face $F_g$ may contain additional $f_i$'s. Recall that the design matrix $A$ given in \eqref{HXtilde} has rows $(1,f_i^t)$. Let $B$ be the $|I|\times |J|$ matrix  with rows $f_i^t, i\in I$. Let $B_+$ be the matrix obtained from $B$ by keeping the rows indexed by $I_+$ only, and let $B_0$ be the matrix obtained by keeping the rows in $I\setminus I_+$ only.
Fienberg and Rinaldo (2012) showed that ${\cal F}\setminus I_+$  could be identified by solving the linear program
\begin{eqnarray}
&&\mbox{Max}\;||Bg||_0\label{prog}\\
&&\mbox{s.t.}\;B_+g=0\nonumber\\
&&\;\;B_0g\geq 0,\nonumber
\end{eqnarray}
where $||.||_0$ is the zero norm in $R^J$. This is  a non-convex program which is, however, easily solved by a sequence of linear programming relaxations (see programs (4) and (6) in Fienberg and Rinaldo (2012), supplementary material). We adopted their approach to find the smallest face containing the data vector $t$ whenever the dimension is small enough for the program to run. We found that we could use this program only for models with up to  16 vertices.
\subsection{ Splitting the global model into smaller models}
Since it is difficult or impossible to implement the program \eqref{prog} in high-dimension, we now consider a collection of smaller models  and we show that the combination of these smaller models yields information on the global model. 

Suppose that $X$ follows a model  ${ M}$ Markov with respect to the undirected graph $G=(V,E)$. Let $A\subset V$ and let $G_A$ be the graph induced by $A$. Let ${M}_A$ be the model Markov with respect to $G_A$; this is also a multinomial model. The generating set ${\cal D}_A$ of ${ M}_A$ is a subset of ${\cal D}$. Let 
 \be \label{jA} J_A=\{ j\in J,\ \ S(j)\in \mathcal{D}_A\}.\ee 
 Then $t_{J_A}=(t(j)=n(j_{S(j)}), \;j\in J_A)$ is the canonical data vector of ${ M}_A$. Let $C_A$ be the marginal cone of ${ M}_A$. Clearly $C_A$ is generated by $f_{i_A, J_A},\;i_A\in I_A$ where $f^t_{i_A,J_A}=(f_{ij}, j\in J_A)$ is made up of the components of $f_i, i\in I$ that are indexed by $J_A$. We may have $f_{i_A,J_A}$ coming from different $f_i$'s but clearly we  keep only one copy.
The following lemma shows  how to extend a face of $C_A$ into a face of $C$.
\begin{lemma}
\label{extend}
For $g_A\in R^{J_A}$, let $F_{g_A}=\{x\in C_A:\;\langle g_A,x\rangle=0\}$ be a face of $C_A$ containing $t_{J_A}$.
Let $g_1=(g_A,0\ldots,0)\in R^J$  be the vector of $R^J$ obtained from $g_A$ by setting the remaining variables to $0$.
Then
$$F_{g_1}=\{x\in C:\; \langle g_1, x\rangle=0\}$$
is a face of $C$ containing $t$.
\end{lemma}
\noindent {\bf Proof.}
Let ${\cal F}_A$ denote the index set of the $f_{i_A}$  defining $F_{g_A}$. We have
\begin{eqnarray*}
\langle g_A, f_{i_A,J_A}\rangle&=&0,\;i_A\in {\cal F}_A\\
\langle g_A, f_{i_A,J_A}\rangle&>&0,\;i_A\not \in {\cal F}_A\\
\langle g_A, t_{J_A}\rangle&=&0.
\end{eqnarray*}
Writing $i\in I$ as $i=(i_A,i_{A^c})$ and $f_i=(f_{i,J_A}, f_{i, J_{A^c}})$, we have
\begin{eqnarray*}
\langle g_1, f_{i}\rangle&=&\langle g_A, f_{i,J_A}\rangle+\langle 0, f_{i, J_{A^c}}\rangle=0,\;i\;\mbox{such that}\; i_A\in {\cal F}_A\\
\langle g_1, f_{i}\rangle&=&\langle g_A, f_{i,J_A}\rangle+\langle 0, f_{i, J_{A^c}}\rangle>0,\;i\;\mbox{such that}\;i\not \in {\cal F}_A\\
\langle g_1, t\rangle&=&\langle g_A, t_{J_A}\rangle+\langle 0, t_{J_{A^c}}\rangle=0.
\end{eqnarray*}
It follows immediately that $F_{g_1}$ is a face of $C$ containing $t$ and it is defined by $f_i, \;i_A\in {\cal F}_A$.
$\square$

\noindent  We now use this lemma for  a finite number of smaller models and in doing so,  obtain an even smaller face containing the data vector $t$. Let ${ M}$ be the model Markov with respect to $G=(V,E)$. Let $G_{A_l},\;l=1\ldots,k$  be a collection of subgraphs induced by subsets $A_l,\;l=1,\ldots,k$  of $V$. Let ${\cal D}_{A_l}$ be the generating set   for the model  ${M}_{A_l}$  Markov with respect to $G_{A_l}$ and let $t_{A_l}$ be its canonical statistic. 
%We note that $t_{J_{A_l}}$ is a sub vector of the canonical statistic vector $t$ of the global model.
\begin{theorem}
\label{main1}
 Let $A_l,\;l=1,\ldots,k$ be  a collection of subsets of $V$.
 % such that ${\cal D}=\cup_{l=1}^k  {\cal D}_{A_l}$. 
 Let $g_{A_l}$  define  faces of $C_{A_l}$  containing $t_{J_{A_l}},\;l=1,\ldots,k$ respectively, let $g_l$ be the vectors of $R^J$ obtained by completing $g_{A_l}$  with zeros and let $F_l=F_{g_l}$  be the corresponding faces. Then 
  $$g=\sum_{l=1}^k g_l$$
 defines a face $\cap_{l=1}^kF_l$ of $C$ containing $t$.
\end{theorem}
\noindent {\bf Proof.}
We give the proof for $k=2$. If $f_i\in F_1\cap F_2$, we have that $\langle f_i, g_1\rangle=\langle f_i, g_2\rangle=0$ and therefore $\langle f_i, g\rangle=0.$ Moreover, if $f_i\not \in F_1\cap F_2$, either $\langle f_i, g_1\rangle>0$ or $\langle f_i, g_2\rangle>0$, and therefore $\langle f_i, g\rangle>0.$ It follows that $F_g=\{x\in C:\langle g, x\rangle=0\}$ is a face of $C$. Moreover,
%{\bf **}
$\langle g,t\rangle=\langle g_1,t\rangle+\langle g_2,t\rangle=0$ and therefore $t\in F_g$ which clearly is equal to $F_1\cap F_2$.
$\square$
\begin{remark}
In practice, when applying the theorem above, we choose the collection $A_l$
% {\bf **} 
 so that   ${\cal D}=\cup_{l=1}^k  {\cal D}_{A_l}$ with  $A_l, l=1,\ldots,k$ as large as possible but small enough  that we can compute the smallest face containing $t_{J_{A_l}}$. In  experiments similar to that presented in Section 3.4 below, we nearly always found that $\cap_{l=1}^kF_l$ was the smallest face containing $t$. 
 \end{remark}
%\subsection{A counterexample}
%We have just shown that if $t$ belongs to the intersection of faces obtained from  smaller models as in Lemma \ref{extend}, the intersection  of these  faces is a face of the global model. In the supplementary file, we give a counterexample showing that the smallest face containing the data vector need not be included in the intersection of faces of the type $F_l$ issued from smaller models of the type ${ M}_{A_l},\;l=1,\ldots,K.$

\subsection{A numerical experiment}
We illustrate our results with a  numerical experiment.
We consider the graphical model with $G$ equal to the four-neighbour $4\times 4$ lattice with binary data so that $|I|=2^{16}$ and $|J|=40$. We generated the data vector
$$t=[1 \    2  \   7  \   5  \   4   \  5    \ 2  \   7 \    7   \  3   \  2   \ 10\  2 \    8  \   1 \    5 \    0  \   0 \    1\     1\     4 \    1 \    4  \ 1\  3 \   1  \   2  \   2  \   0  \   7  \  3  \   1  \   1   \  2  \   2    \ 1\ 5 \ 2 \ 0 \ 0].$$
We let $A_i, i=1,2,3$ be the subsets of $V$ comprising the vertices in the $i$-th and $(i+1)$-th row of the lattice. We find that the face $F_{A_1}$ of $C_{A_1}$ containing $t_{J_{A_1}}$ is of dimension 15 and the expanded face $F_1$ is therefore of dimension $15+22=37$. The corresponding $g_1$ is
\[
 g_1=10\;\Big(0
    , \ 0
       ,\   0
    , \ 0
   ,\ 0
      ,\    0
   ,\ 1
   , \ 0
   ,\ 1
   ,\ 1
   ,\ 0
    ,\ 0
    ,\ 0
   ,\ 0
   ,\ 0
  ,\  0
    ,\ 0,\ 
  -1\Big).
 \]
The face $F_{A_2}$ is of dimension 13 while $F_2$ is of dimension $13+22=35$. The corresponding vector is 
\[g_2=10\;\Big(
0,\ 0,\ 	1, \ 1, \ 0,\ 	1,\ 	1,\	 0,\ 0,\ 0, \ 	0,\ 0, \ -1,\ 1,\ -1,\ -1, \ 0, \ 	-1 \Big)
\] The face $F_{A_3}$ is of dimension 11 and $F_3$ is of dimension $11+22=33$ with
\[
g_3=10\;\Big(0	 ,\ 1,\ 	1,\ 	0,\ 	1,\ 	0,\ 	1,\ 	1,\ 	-1,\ 	0,\ 	0,\ 	0,\ 	-1,\ 	-1,\ 	-1,\ 	-1,\ 	1,\ 	1\Big).
\] In this particular case, after computing the facial set of $F$, the smallest face of the global model containing $t$, we find that actually $F=F_1\cap F_2\cap F_3$ which is of dimension 28.

\section{The  conditional and marginal composite maximum likelihood estimators}
 When the dimension of the discrete graphical model is large, computing the maximum likelihood estimate of $\theta$  in \eqref{ftheta}  is challenging, if not impossible. As mentioned in the introduction, a recent approach to this problem has  been local with the use of a composite likelihood which is equal to the product, over all vertices $v\in V$, of the local conditional likelihood for $X_v$ given $X_{\N_v}$ where $\N_v$ denotes the set of neighbours of $v$ in $G$. Recently, for  Gaussian high-dimensional graphical models, Wiesel \& Hero (2012) and Meng \& al. (2013, 2014) worked with a different composite likelihood which is the product,  over all vertices $v\in V$, of local marginal likelihoods. In this section, we will first recall the definition of the conditional composite likelihood estimate, then extend the marginal composite likelihood to discrete graphical models and finally  show that the  maximum  likelihood estimates obtained from these two types, conditional and marginal,  of local models are in fact identical and thus the composite likelihood obtained by consensus from these two types of likelihood are equal. Since the computational complexity of the marginal computations is exponential in the number of vertices in the neighbourhood of $v$ while the conditional computations are linear in this number, there is no advantage in working with marginal composite likelihoods.

\subsection{The conditional composite likelihood function}
We first define the standard conditional composite likelihood function.
For $i=(i_v, v\in V)$, let $i^{(1)},\ldots,i^{(N)}$ be a sample of size $N$ from the distribution of $X$ Markov with respect to $G$.  We recall that the global likelihood function is
\begin{equation}
\label{global}
L(\theta)\propto \prod_{k=1}^Np(X_v=i_v^{(k)}, v\in V)=\exp \{\langle \theta, t\rangle -Nk(\theta)\}
\end{equation}
where $k(\theta)$ is as in \eqref{k}. 

For a given vertex $v\in V$, let $\N_v$ the set of neighbours of $v$ in the given graph $G$. The composite likelihood function based on the local conditional distribution of $X_v$ given $X_{V\setminus\{v\}}$ or equivalently, due to the Markov property,  the conditional distribution of $X_v$ given its neighbours $X_{\N_v}$ is $L^{PS}(\theta)=\prod_{v\in V}L^{PS_{v}}(\theta)$ where
\begin{equation}
\label{compcond}
L^{PS_v}(\theta)=\prod_{k=1}^Np(X_v=i_v^{(k)}|X_{\N_v}=i_{\N_v}^{(k)}; \theta)
\end{equation}
and  the superscript $PS$ stands for "pseudo-likelihood",  the name often given to the conditional composite lilelihood (Besag, 1974).
As given by \eqref{pi}, for a given cell $i$, we have
\begin{eqnarray*}
\log p(i)&=&\log p(X_v=i_v, v\in V)=\theta_0+\sum_{j\tl i}\theta_j\\
&=& \theta_0+\sum_{j\tl i,\;S(j)\subseteq {v\cup \N_v},S(j)\not \subseteq \N_v}\theta_j +\sum_{j\tl i,\;S(j)\subseteq { \N_v}}\theta_j+\sum_{j\tl i,\;S(j)\not \subseteq {v\cup \N_v}}\theta_j    
\end{eqnarray*}
The set $J$ is as defined in \eqref{j} for the global model. Let 
$$J^{PS_v}=\{j\in J\mid S(j)\subseteq {v\cup \N_v},S(j)\not \subseteq \N_v\}=\{j\in J\mid v\in S(j)\}.$$
Then for $i_v\not =0$, we have
\begin{align}
&&p(X_v=i_v|\;X_{\N_v}=i_{\N_v})=p(X_v=i_v|\;X_{V\setminus \{v\}}=i_{V\setminus \{v\}})=\frac{p(X_{V}=i_V)}{p(X_{V\setminus \{v\}}=i_{V\setminus \{v\}})}\nonumber\\
&=&\frac{e^{\theta_0+\sum_{j\tl i,\;j\in J^{PS_v}}\theta_j +\sum_{j\tl i,\;S(j)\subseteq { \N_v}}\theta_j+\sum_{j\tl i,\;S(j)\not \subseteq {v\cup \N_v}}\theta_j  }
}
{\sum_{k\in I|\;k_{V\setminus \{v\}}=i_{V\setminus \{v\}}}\Big(e^{\theta_0+\sum_{j\tl k,\;j\in J^{PS_v}}\theta_j +\sum_{j\tl k,\;S(j)\subseteq { \N_v}}\theta_j+\sum_{j\tl k,\;S(j)\not \subseteq {v\cup \N_v}}\theta_j  }\Big)
}\nonumber\\
&=&\frac{e^{\sum_{j\tl i,\;j\in J^{PS_v}}\theta_j }
}
{1+\sum_{k\in I|\;k_{V\setminus \{v\}}=i_{V\setminus \{v\}},\; k_v\not =0}e^{\sum_{j\tl k,\;j\in J^{PS_v}}\theta_j }
}\label{4.3}
\end{align}
and 
\begin{eqnarray}
p(X_v=0|\;X_{V\setminus \{v\}}&=&i_{V\setminus \{v\}})=\frac{1}{1+\sum_{k\in I|\;k_{V\setminus \{v\}}=i_{V\setminus \{v\}},\; k_v\not =0}e^{\sum_{j\tl k,\;j\in J^{PS_v}}\theta_j }\label{4.4}
}
\end{eqnarray}
Equality \eqref{4.3} is due to the fact that  the set of $j\in J$ such that $j\tl k,\; S(j)\not \subseteq {v\cup \N_v},$ is the same whether $k_v=i_v$ or $k_v\not =i_v$ and therefore the term $e^{\theta_0+\sum_{j\tl k,\;S(j)\not \subseteq k_{v\cup \N_v}}\theta_j }$ cancels out at the numerator and the denominator. The same goes for the set of $j\in J$ such that $j\tl k,\; S(j)\subseteq \N_v$. 

\begin{remark}
\label{tvps}
%We first note that the $1$ at the denominator of the probabilities above corresponds to the term where $i_v=0$. 
%\begin{enumerate}
%\item[(a)]We  note that the only $j$'s appearing in the formula above are such that their support is in $v\cup \N_v$ but not in $\N_v$. So writing $j\tl i$ is equivalent to $j_{v\cup \N_v}\tl i_{v\cup \N_v}$ where it is understood that if $S(j)\subseteq S(i_{v\cup \N_v})$, then $j_{v\cup \N_v}=j$ where $j_{v\cup \N_v}$ is padded with 0's in $V\setminus (v\cup \N_v)$.
%\item[(a)] 
In the equation above, we worked with $p(X_v|X_{V\setminus \{v\}})$ rather than with $P(X_v|X_{\N_v})$,  though the two are equal, in order to emphasize that the parameter
\begin{equation}
\label{thetacond}
\theta^{PS_v}=(\theta_j, \;j\in J^{PS_v}),\;\;v\in V
\end{equation}
 of the $v$-th component $L^{PS_v}$ of conditional composite distribution is a subvector of $\theta$, the parameter of the global likelihood function.
\end{remark}
We now define the two-hop conditional composite likelihood function.
\begin{definition}
For a given $v\in V$, we will say that $\M_v$ is a one-hop neighbourhood of $v$ if it comprises $v$ and its immediate neighbours in $G$, i.e. if $\M_v=\{v\}\cup \N_v$. We will say that $\M_v$ is a two-hop neighbourhood if it comprises $v$, its immediate neighbours and the neighbours of the immediate neighbours in $G$. We use the notation
$$\N_{2v}=\M_v\setminus \Big(\{v\}\cup \N_v\Big)$$
to denote the set of neighbours of the neighbours of $v$. For simplicity of notation, we will denote both the one-hop and two-hop neighbourhoods by $\M_v$. 
\end{definition}
The two-hop conditional composite likelihood function is $L^{PS_2}(\theta)=\prod_{v\in V}L^{PS_{2,v}}(\theta)$
\begin{equation}
\label{compcond2}
L^{PS_{2,v}}(\theta)=\prod_{k=1}^Np(X_v=i_v^{(k)}, X_{\N_v}=i_{\N_{v}}^{(k)}|X_{\N_{2v}}=i_{\N_{2v}}^{(k)}).
\end{equation}
The expression of 
$p(X_v=i_v^{(k)}, X_{\N_v}=i_{\N_{v}}^{(k)}|X_{\N_{2v}}=i_{\N_{2v}}^{(k)})$ is the same  as \eqref{4.3} and \eqref{4.4} but with $J^{PS_v}$ replaced by $J^{PS_{2,v}}$ where
$$J^{PS_{2,v}}=\{j\in J\mid S(j)\subseteq { \M_v},S(j)\not \subseteq \N_{2v}\}.$$
In a parallel way to Remark \ref{tvps}, we note that
$$\theta^{PS_{2,v}}=\{\theta_j, j\in J^{PS_{2,v}}\}$$
is a subvector of $\theta=(\theta_j, j\in J)$, the argument of the global likelihood function.
\subsection{The marginal composite likelihood}
Let $\M_v$ be the one-hop or two-hop neighbourhood of $v$. The marginal composite likelihood is the product
\begin{equation}
\label{compmarg}
L^{\M}(\theta)=\prod_{v\in V}\prod_{k=1}^Np(X_{\M_v}=i_{\M_v}^{(k)})=\prod_{v\in V}L^{\M_v}(\theta).
\end{equation}
where 
%$\tilde{\theta}$ is the parameter of the composite marginal likelihood and 
$L^{\M_v}(\theta)=\prod_{k=1}^Np(X_{\M_v}=i_{\M_v}^{(k)})$.
%We first consider  the $\M_v$-marginal model  derived from \eqref{ftheta} with cell probabilities $p(i_{\M_v})=p(X_v=i_v,X_{\N_v}=i_{\N_v})$.
The $\M_v$-marginal model   is clearly multinomial  and the corresponding data  can be read in the $\M_v$-marginal contingency table obtained from the full table. 
%Since in general, the model is not collapsible onto the graph induced from $G$ by $\M_v$, we will first have to derive the generating set  ${\cal D}^{\M_v}$ of the marginal model.
%Let $J^{\M_v}$ be the $J$-set indexing the  nonzero canonical parameters of the $\M_v$-marginal model, that is
%$$J^{\M_v}=\{i\in \I_{\M_v}:\; S(i)\in {\cal D}^{\M_v}\}.$$
%The corresponding canonical  parameter, i.e. the argument of $L^{\M_v}$,  is therefore
%$$\theta^{\M_v}=(\theta_j^{\M_v}, \;j\in J^{\M_v}).$$
%The marginal distribution of $X_{\M_v}$ has density of the form
The density of the $\M_v$-marginal multinomial distribution is of the general exponential form
\begin{equation}
\label{marginal}
f(t^{\M_v};\;\theta^{\M_v})=\exp\{ \langle t^{\M_v}, \theta^{\M_v}\rangle -Nk^{\M_v}(\theta^{\M_v})\}
\end{equation}
where $ t^{\M_v},\;\theta^{\M_v}$ and $k^{\M_v}$ are respectively the $\M_v$-marginal canonical statistic, canonical parameter and  cumulant generating function. 

In order to identify the $\M_v$-marginal model, we first establish the relationship between $\theta$ and $\theta^{\M_v}$. In the sequel, the symbol $j$ will be understood to be an element of $I_{\M_v}$ when used in the notation $\theta_j^{\M_v}$ while it will be understood to be the element of $J$ obtained by padding it with entries $j_{V\setminus \M_v}=0$  when used in the  notation $\theta_j$. 
%It will also be understood that $\theta_j=0$ if $j\not\in J$ and $\theta_j^{\M_v}=0$ when $j\not \in J^{\M_v}$.
We now give the general relationship between the parameters of the overall model and those of the $\M_v$-marginal model. Proofs are given in the Appendix.
\begin{lemma}
\label{one}
Let $\M_v$ be the one-hop or two-hop neighbourhood of $v\in V$. For $j\in J, S(j)\subset \M_v$, the parameter $\theta_j$ of the overall model and the parameter $\theta_j^{\M_v}$ of the marginal model are linked by the following:
\begin{align}
\label{main}
 \theta_{j}^{\M_v}
 &=&\theta_{j}+\sum_{j'\;|\;j'\tl_0 j}(-1)^{|S(j)-S(j')|}\log \Big(1+\sum_{i\in \I,\; i_{\M_v}=j'}\exp \sum_{{k\;|\;k\tl i \atop k\not \tl j'} }\theta_{k} \Big)\;.
\end{align}
\end{lemma}

We now want to identify which of the marginal parameters are equal to the corresponding overall parameter and in particular which marginal parameters are equal to $0$ when the global parameter is equal to zero. Let $\M_v^c$ denote the complement of $\M_v$ in $V$. We define the buffer set  at $v$ as follows:
\begin{equation}
\label{bv}
\B_v=\{w\in \M_v\;|\;\exists w'\in \M^c_v\;\mbox{with}\;(w,w')\in E\}.
\end{equation}
We have the following result.
\begin{lemma}\label{two}
Let $\M_v$ be the one-hop or two-hop neighbourhood of $v\in V$. For $j\in J, S(j)\subset \M_v$ the following holds:
\begin{enumerate}
\item[(1.)] if $S(j)\not \subset\B_v$, then $\theta^{\M_v}_{j}=\theta_j$,
\item[(2.)] if $S(j)\subset\B_v$, then in general $\theta^{\M_v}_j\not =\theta_j$, and \eqref{main} holds.
\end{enumerate}
Moreover, for $i\in I, S(i)\subset \M_v$, 
\begin{enumerate}
\item[(3.)] If $S(i)\not \subset \B_v$, then $\theta_i^{\M_v}=0$ whenever $\theta_i=0.$
\end{enumerate}
%In particular this implies that  
%$$\theta_j=0\;\mbox{and}\;S(j)\not \subset\B_v \Rightarrow \theta^{\M_v}_{j}=0\;\mbox{i.e.}\;j\not \in J^{\M_v}$$ 
%while if $\theta_j=0$ but $S(j) \subset\B_v$, then $j$ might belong to $J^{\M_v}$.
\end{lemma}

\noindent From the lemma above, we see that, for $j\in J$ such that $S(j)\subset \M_v, S(j)\not \subset \B_v$, the corresponding global and $\M_v$-marginal loglinear parameters are equal. We see also that for $i\in I$ such that $ S(i)\in \M_v, S(i)\not \subset \B_v$, if  the loglinear parameter is zero in the global model, it remains zero in the $\M_v$-marginal model.

\subsection{A convex relaxation of the local marginal optimization problems}
It is clear from \eqref{main} that even though maximizing the marginal likelihood from \eqref{marginal}   is  convex in $\theta^{\M_v}$, it is not convex in $\theta$. We would therefore like to replace the problem of maximizing \eqref{marginal} non convex  in $\theta$ by a convex relaxation problem. 
We know from  {\it (1.)} of Lemma \ref{two} that $\theta_j^{\M_v}=\theta_j$ for $j$ in the set
$\{j\in J:\; S(j)\subset \M_v, \;S(j)\not \subset \B_v\}\;.$

We also know from {\it (3.)} of Lemma \ref{two}  that if the  global model parameter $\theta_i, S(i)\subset \M_v, S(i)\not \subset \B_v$ is equal to zero, then $\theta_i^{\M_v}$ is also equal to zero. Following what has been done for Gaussian graphical models, it is then natural to consider the following graphical model relaxation of the $\M_v$-marginal model.
 
 Let  $\M_{l,v}$ index the relaxed hierarchical loglinear model  obtained from the $\M_v$-marginal model by keeping interactions given by edges with at least one endpoint in $\M_v\setminus \B_v$ and only those edges, and all interactions in the power set $2^{\B_v}$. The index $l$ takes values $l=1$ or $l=2$ when $\M_v$ is respectively the one-hop or two-hop neighbourhood of $v$.
 The $J$-set of this local model is
\begin{equation}
\label{jm1v}
J^{\M_{l,v}}=\{j\in J\mid S(j)\subset \M_v, S(j)\not \subset \B_v\}\cup \{ i\in I\mid S(i)\subset \B_v\}\;.
\end{equation}

Let $p^{\M_{l,v}}(X_{\M_{v}})$ denote the marginal probablity of $X_{\M_v}$ in the $\M_{l,v}$-marginal model. The local estimates of $\theta_j, j\in \{j\in J|\; S(j)\subset \M_v,\; S(j)\not \subset \B_v\}$ are obtained by maximizing the $\M_{l,v}$-marginal loglikelihood 
\begin{eqnarray}
\label{relax}
L^{\M_{l,v}}(\theta)=\prod_{k=1}^N p^{\M_{l,v}}(X_{\M_{v}}=i^{(k)}_{\M_{v}})=\exp\{ \langle \theta^{\M_{l,v}},t^{\M_{l,v}}\rangle -Nk^{\M_{l,v}}(\theta^{\M_{l,v}})\}
\end{eqnarray} 
which is a convex maximization problem in 
$$\theta^{\M_{l,v}}=(\theta_j, j\in J^{\M_{l,v}}).$$

%In practice, we keep the estimates of $\theta_j$ for $S(j)$ made up of $v$ and its immediate neighbours whether $\M_{l,v}$ denotes the one-hop or two-hop neighbourhood. However in principle we could keep all $\theta_j, j\in J^{\M_{l,v}}$ with $S(j)\not \subset \B_v$.
 At this  point, we need to make two important remarks. 
 \begin{remark}
 The vector $\theta^{PS_v}$ defined in \eqref{thetacond} is a subvector of $\theta^{\M_{l,v}}$.
%In particular for $l=1$, 
% $$\theta^{\M_{1,v}}=(\theta^{PS_v}, (\theta_i, i\in i, S(i)\subset \B_v)).$$
 Therefore maximizing \eqref{relax} for either $l=$ or $l=2$ will yield an estimate of $\theta^{PS_v}$.
 \end{remark}
 \begin{remark}
 The $\M_{l,v},\;l=1,2$-marginal model is a hierarchical loglinear model but not necessarily a graphical model. For example, if we consider a four-neighbour lattice and a given vertex $v_0$ and its four neighbours that we will call $1,2,3,4$ for now, then the generating set of the relaxed $\M_{1,v_0}$-marginal model is
 $${\cal D}^{\M_{1,v_0}}=\{(v_0,1),  (v_0,2), (v_0,3), (v_0,4), (1,2,3,4)\}.$$
 This is not a discrete graphical model since a graphical model would also include the interactions $(v_0,1,2),(v_0,2,3),(v_0,3,4),(v_0,1,4),(v_0,1,2,3,4).$
 It was therefore crucial to set up our problem, as we did it in Section 2, within the framework of hierarchical loglinear models rather than the more restrictive class of discrete graphical models.
 \end{remark}
 \vspace{2mm}

\subsection{ The  composite maximum likelihood estimates}
To obtain the composite mle of the global loglinear parameter $\theta$, we do the following. First, for each $v\in V$, we compute the maximum likelihood estimates from $L^{PS_v}$   as in \eqref{compcond} (in the conditional composite likelihood case) or for $L^{\M_{l,v}}$  in \eqref{relax} (in the marginal composite likelihood case).
Second, in each case, we retain only the estimate of $\theta^{PS_v}.$ Third, the global maximal composite likelihood estimate of each $\theta_j, j\in J$ is derived either by simple averaging of the different estimates of $\theta_j^{PS_v}$ or $\theta^{\M_{l,v}}_j, j\in J^{PS_v}$ obtained from the various local models 
or by more sophisticated ways such as described in  Liu and Ihler (2012). This includes linear consensus, maximum consensus or ADMM. If one uses the two-hop neighbourhood, then the accuracy is such that usually simple averaging is sufficient to obtain very good accuracy.

 If follows immediately that if we can prove that the mle of $\theta^{PS_v}$ obtained from $L^{PS_v}$ and from $L^{M_{l,v}}$ are identical, then the maximal composite likelihood estimates of $\theta$ obtained from the conditional or the marginal composite likelihood by consensus will be the same. In the next subsection, we show that this is indeed the case. 
%$PS$ component $\hat{\theta}^{\M_{1,v}}_{PS}$ of the local mle $\hat{\theta}^{\M_{1,v}}$ of $L^{\M_{l,v}}$ is equal to the mle $\hat{\theta}^{PS}$ of $L^{PS}$ 
%and similarly for $\hat{\theta}^{\M_{2,v}}_{PS}=\hat{\theta}^{PS_2}$.

\subsection{Equality of the maximal conditional and marginal composite likelihood estimate }
Let $\hat{\theta}^{\M_{l,v}}, l=1,2$ denote the maximum likelihood estimate of $\theta^{\M_{l,v}}$ obtained from the local likelihood \eqref{relax}.
\begin{theorem}
\label{equal}
The $PS$ component of $\hat{\theta}^{\M_{1,v}}$,i.e. $(\hat{\theta}_j^{\M_{1,v}}, j\in J^{PS_v})$ is equal to the maximum likelihood estimate of ${\theta}^{PS_v}$ obtained from the local conditional likelihood \eqref{compcond}.

\noindent Similarly, The $2PS$ component of $\hat{\theta}^{\M_{2,v}}$,i.e. $(\hat{\theta}_j^{\M_{2,v}}, j\in J^{PS_{2,v}})$ is equal to the maximum likelihood estimate of ${\theta}^{PS_{2,v}}$ obtained from the local conditional likelihood \eqref{compcond2}.
\end{theorem}
The proof is given in the Appendix.
%We conclude immediately from this theorem that, if the composite likelihood estimate is obtained by consensus, there is no point in computing it from local marginal composite likelihoods \eqref{relax} since in this case the computational complexity  is exponential in the number of vertices while for local conditional likeilhoods the computational complexity is linear in the number of vertices in $\M_v$.

\begin{figure}
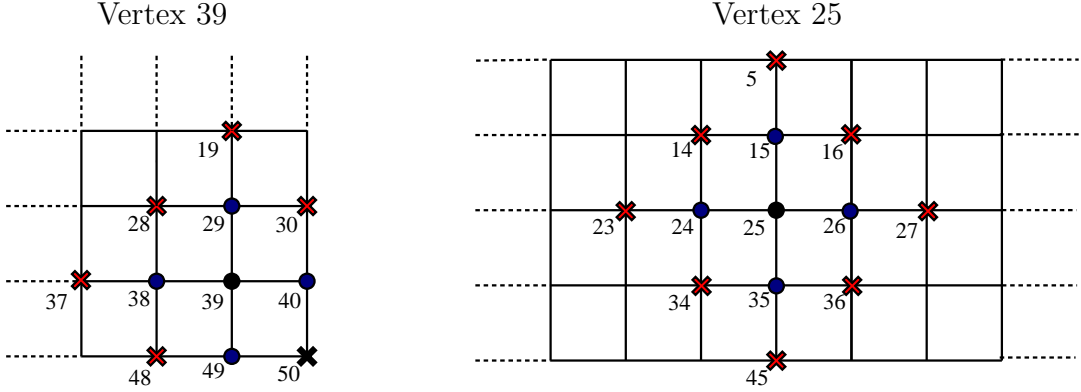

\begin{center}
\begin{tabular}{ccccc}
Vertex 39 &&&& Vertex 25 \\
\scalebox{1}{\includegraphics{39.pdf}}&&&&\scalebox{1}{\includegraphics{25.pdf}}\\
\end{tabular}
\caption{Two vertices in a $5\times 10$ lattice: Theorem \ref{equal} applies for vertex 25 while it does not apply for vertex 39.}
\label{fig1}
\end{center}
\end{figure}

At this point, we ought to make an important observation. In the case of the two-hop marginal likelihood, it may happen that the buffer $\B_v$ is no longer equal to $\N_{2v}$. For example, if we consider a four-neighbour $5\times 10$ lattice, the vertex $39$ is such that $\N_v^2=\{19,28,30,37,48,50\}$ while $\B_v=\N_{2v}\setminus \{50\}$. The argument in the proof of Theorem \ref{equal} for $j$ such that $S(j)\not \subset \N_{2v}$ then breaks down since in the $\M_{2,v}$-marginal model,  some cells such as $i_{\M_v}=(i_{30}=1,i_{50}=1,0_{\M_v\setminus \{30,50\}})$ with support in $\N_{2v}$ no longer have a complete support. This situation is illustrated in Figure \ref{fig1} where for the sake of comparison, we also draw a vertex for which $\N_{2v}=\B_v$ and Theorem \ref{equal} applies..

In Tables \ref{tab1} and  \ref{tab2}, we give the numerical values of the maximum likelihood estimate $\theta_j, j\in J^{\M_{2,v}}$ obtained by the four local model $PS, PS_2, \M_{1,v}$ and $\M_{2,v}$ for $j$ such that $j\in J^{PS_{25}}$ and for $j$ such that $j\in J^{PS_{39}}$, respectively. We see that in the first case, the values of $\hat{\theta}_j$ obtained from the  local likelihoods $l^{PS_{25}}$ and $l^{\M_{1,25}}$ are identical and similarly for those obtained from $l^{PS_{2,25}}$ and $l^{\M_{2,25}}$,
while in the second case, the values obtained through the $PS_2$ and $\M_{2,v}$ models are slightly different. The values obtained from the $PS$ and $\M_{1,v}$ models are identical since then $\B_v=\N_v$ and the proof of Theorem \ref{equal} does not break down.

\begin{table}[h!]
  \begin{center}
    \begin{tabular}{| c| c| c |c|c|c|}
    \hline
    Models & $\hat{\t}_{25}$ & $\hat{\t}_{15,25}$ & $\hat{\t}_{24,25}$ & $\hat{\t}_{25,26}$ & $\hat{\t}_{25,35}$\\
    \hline
    $\M_{1,v}$ & -0.0536 & 0.5914 &  -0.4808 &  -0.8314 &  -0.8461  \\ 
    \hline
     $\M_{2,v}$ & -0.0779 & 0.5221 & -0.5310 & -0.7274 & -0.7459  \\
     \hline
      $(v,PS)$ & -0.0536 & 0.5914 &  -0.4808 &  -0.8314 &  -0.8461  \\ 
      \hline
       $(v,2PS)$ & -0.0779 & 0.5221 & -0.5310 & -0.7274 & -0.7459  \\  
    \hline
    \end{tabular}
  \end{center}
  \caption{The composite mle of some $\theta_j, j\in J^{25,PS}$  in the $5 \times 10$ lattice}
  \label{tab1}
\end{table}

\begin{table}[h!]
  \begin{center}
    \begin{tabular}{| c| c| c |c|c|c|}
    \hline
    Models & $\hat{\t}_{39}$ & $\hat{\t}_{29,39}$ & $\hat{\t}_{38,39}$ & $\hat{\t}_{39,40}$ & $\hat{\t}_{39,49}$\\
    \hline
    $\M_{1,v}$ & -1.0799 &  -0.3306 &  -0.3647 &  -0.5791 &  1.1749  \\ 
    \hline
     $\M_{2,v}$ & -1.0386 & -0.3519 & -0.5020 & -0.5445 & 1.1946  \\
     \hline
      $(v,PS)$ & -1.0799 &  -0.3306 &  -0.3647 &  -0.5791 &  1.1749   \\ 
      \hline
       $(v,2PS)$ & -1.0381 & -0.3531 & -0.5019 &  -0.5448 &  1.1947  \\  
    \hline
    \end{tabular}
  \end{center}
  \caption{The composite mle of some $\theta_j, j\in J^{39,PS}$   in the $5 \times 10$ lattice}
  \label{tab2}
\end{table}

\begin{remark}
The equality of the estimates holds also for the marginal estimates obtained by Mizrahi et al. (2014b) if, for $q$ a clique of $G$ and $v\in q\subset {\cal A}_q$, satisfying the strong LAP condition with respect to ${\cal A}_q$, we retain only the parameters $\theta_j, j\in J^{PS_v}\cap q.$ We also note that Theorem 9 in that paper may not be verified in some cases. For example, take vertex 7 in a $3\times 3 $ lattice numbered from left to right starting with the top row, take $q=\{7,8\}$ as the clique of interest. Then ${\cal A}_q=\{4,7,8\}$ satisfies the strong LAP condition but $\theta_8$ in the ${\cal A}_q$-marginal model cannot be equal to $\theta_8$ in the joint model  as our Lemma \ref{two} shows.
\end{remark}
\subsection{Existence of the mle and the composite likelihood estimates}
To finish this section, we describe a numerical experiment illustrating the impact of the non existence of the global mle on the computations of the various composite maximum likelihood estimates.

 For each sample size $N=40,60$ and $ 80$, we consider two experiments, one where the data point $t=(t_j,j\in J)=(n(j_{S(j)}), j\in J)$ belongs to a face of the global model and one where it does not. For each experiment, we compute five estimates, the global mle, and the four composite likelihood estimates based on  $\hat{\theta}^{\M_{1,v}}, \hat{\theta}^{PS_v}, \hat{\theta}^{\M_{2,v}}, \hat{\theta}^{PS_{2,v}}, v\in V.$ 
For each one of the five estimates and for each experiment, we report the relative mean square error (MSE)
$$\frac{||\hat{\theta}-\theta^*||^2}{||\theta^*||^2}$$
where $\theta^*$ denotes the true value of the  parameter. 
The results are given in Table \ref{mse}

 \begin{table}[h!]
  \begin{center}
    \begin{tabular}{|c|c|c|c|c|c|c|c|c|c|c|}
    \hline
    Sample size &40(on face)&40 (not on face) & 60(on face) & 60(not on face) & 80(on face) & 80(not on face)  \\
    \hline
    Global MLE & 102.72 & 3.1270 & 39.68& 2.2620 & 27.12 &1.5717\\
    \hline
    $\M_1$ MLE & 134.94 & 3.6335 & 66.00& 2.3214 & 43.46  & 1.6454\\
    \hline
    $PS_1$ MLE & 127.64 & 3.6335 & 32.63 & 2.3214 & 24.34 & 1.6454 \\
    \hline
    $\M_2$ MLE &340.81 & 3.1328 & 65.01 & 2.2700 &  42.67 & 1.5728 \\
    \hline
    $PS_2$ MLE &84.52 & 3.1320 & 44.55 & 2.2709 & 29.76 & 1.5727 \\
    \hline
  \end{tabular}
  \end{center}
  \label{mse}
  \caption{The relative MSE of different estimates of $\theta$ for the $5 \times 10$ lattice when the mle exists and when it does not.}
  \end{table}

We note that when the global mle does not exist the mean square error for all experiments is much larger than when the mle exists indicating that some of the local mle estimates do not exist. When a local mle does not exist, a routine maximization of the local likelihood may lead to erroneous results. We  have thus illustrated that the numerical results from each local maximization must be examined carefully to detect any potential existence problem.
We note also that the data vector may be on a face of a global model without being on the face of any local model. In this case, clearly, the composite mle of $\theta$ is not a consistent estimate of $\theta_0$.

\section{Asymptotic properties of the maximum composite likelihood estimate}
%Since $\hat{\theta}^{PS_v}=(\hat{\theta}^{\M_{l,v}})_{PS}$ and since the computational complexity for $\hat{\theta}^{\M_{l,v}}$ is greater than for $\hat{\theta}^{PS_v}$,  there is no reason to use composite marginal likelihood rather than conditional composite likelihood to approximate the global maximum likelihood. 
The asymptotic properties of the maximum composite likelihood estimate when $p$ is fixed and $N$ goes to $+\infty$ are well-known (see Jordan and Liang (2009)). In this section, we consider the asymptotic properties of the conditional composite mle (which is also the marginal composite mle) when both $p$ and $N$ go to $+\infty$.  In Theorem \ref{mbound} below, we give its rate of convergence  to the true value $\theta^*$.  In order to compare the behaviour of the composite mle with the global mle, we also give, in Theorem \ref{global}, the rate of convergence of the global mle under the same asymptotic regime. 
It will be convenient to introduce the notation
\[
f_j(x)=\prod_{l \in S(j)} \mathds{1}(x_l=j_l)=\left\{\begin{array}{cc}1&\mbox{if}\;j\tl x\\0&\mbox{otherwise}\end{array},\right .
\]
and to write \eqref{4.3} as
\begin{equation}
p(x_v|x_{N_v})=\frac{\exp \{\sum_{j \in J^{PS_v}}\t_j f_{j}(x_v,x_{N_v})\}}{1+ \sum_{y_v \in I_v \setminus \{0\}} \exp \{\sum_{j \in J^{PS_v}}\t_j f_{j}(y_v,x_{N_v})\}}\;.
\end{equation}
In this section, we work exclusively with $l^{PS_v}(\theta^{PS_v})$. Therefore for simplicity of notation we write $\theta$ for $\theta^{PS_v}$. Also, for convenience, we scale the log likelihood by the factor $\frac{1}{N}$. Then the  $v$-local conditional log-likelihood function  is 
\[
\begin{array}{lcl}
l^{PS_v}(\t)&=&\frac{1}{N}\sum_{n=1}^N \log p(x_v^{(n)}|x_{N_v}^{(n)})\\
&=& \sum_{j \in J^{PS_v}}\t_j \frac{1}{N}\sum_{n=1}^Nf_{j}(x_v^{(n)},x_{N_v}^{(n)}) \\&-&\frac{1}{N}\sum_{n=1}^N \log \{1+ \sum_{y_v \in I_v \setminus \{0\}} \exp \{\sum_{j \in J^{PS_v}}\t_j f_{j}(y_v,x^{(n)}_{N_v})\}\}
\end{array}
\]
The sufficient statistic is $t_j =\frac{1}{N}\sum_{n=1}^Nf_{j}(x_v^{(n)},x_{N_v}^{(n)}).$ 
We write 
\begin{equation}\label{yv}t_{J^{PS_v}}=[t_1,t_2,\cdots,t_{d_v}]
\end{equation}
and 
\[
k^{PS_v}(\t)= \frac{1}{N}\sum_{n=1}^N \log \{1+ \sum_{y_v \in I_v \setminus \{0\}} \exp \{\sum_{j \in J^{PS_v}}\t_j f_{j}(y_v,x^{(n)}_{N_v})\}\}=\frac{1}{N}\sum_{n=1}^N \log Z^{n,v}(\theta)
,\] 
where 
$$Z^{n,v}(\theta)=1+ \sum_{y_v \in I_v \setminus \{0\}} \exp \{\sum_{j \in J^{PS_v}}\t_j f_{j}(y_v,x^{(n)}_{N_v})\}.$$
Then the log -likelihood function is 
\[
l^{PS_v}(\t)=\sum_{j \in J^{PS_v}}\t_j t_j-k^{PS_v}(\t)\;.
\]
Its first derivative is
\begin{align}
\frac{\partial l^{PS_v}(\t)}{\partial \t_k}&=t_k-\frac{\partial k^{PS_v}(\t)}{ \partial \t_k},\nonumber\\
\frac{\partial k^{PS_v}(\t)}{ \partial \t_k}&=\frac{1}{N}\sum_{n=1}^N
\frac{ \exp \{\sum_{j \in J^{PS_v}}\t_j f_{j}(k_v,x^{(n)}_{N_v})\}\}}{ Z^{n,v}(\t)}f_{k}(k_v,x_{N_v}^{(n)})\nonumber
\end{align}
with
\begin{equation}
\label{p}
\frac{ \exp \{\sum_{j \in J^{PS_v}}\t_j f_{j}(k_v,x^{(n)}_{N_v})\}\}}{ Z^{n,v}(\t)}=p(X_v=k_v|x_{N_v}^{(n)})
\end{equation}
We now want to compute $
\frac{\partial^2 l^{PS_v}(\t)}{\partial \t_k \partial \t_l}=-\frac{\partial^2 k^{PS_v}(\t)}{ \partial \t_k \partial \t_l}\;, k,l \in J^{PS_v}.
$ To simplify further our notation, we set
\begin{equation}
\label{ztheta}
z_{y_v}(\theta)=\sum_{j \in J^{PS_v}} \t_j f_j(y_v,x_{N_v}^{(n)}).
\end{equation}

For  $k_v=l_v$, using \eqref{p}, we obtain
\[
\begin{array}{lcl}
\frac{\partial^2k^{PS_v}(\t)}{\partial \t_k \partial \t_l}&=&\frac{1}{N}\sum_{n=1}^N
\Big(\frac{ \exp z_{k_v}(\theta)}{ Z^{n,v}(\t)} -(\frac{ \exp z_{k_v}(\theta)}{ Z^{n,v}(\t)})^2\Big) f_{k}(k_v,x_{N_v}^{(n)})f_{l}(l_v,x_{N_v}^{(n)}) \\ 
&=& \frac{1}{N}\sum_{n=1}^N \Big(p(X_v=k_v|x_{N_v}^{(n)})-p(X_v=k_v|x_{N_v}^{(n)})^2)f_{k}(k_v,x_{N_v}^{(n)})f_{l}(l_v,x_{N_v}^{(n)}\Big)\;.
\end{array}
\]
if $k_v \not = l_v$, then
\[
\begin{array}{lcl}
\frac{\partial^2k^{PS_v}(\t)}{\partial \t_k \partial \t_l}&=&\frac{1}{N}\sum_{n=1}^N
-\frac{ \exp z_{k_v}(\theta) \exp z_{l_v}(\theta)}{ (Z^{n,v}(\t))^2}  f_{k}(k_v,x_{N_v}^{(n)})f_{l}(l_v,x_{N_v}^{(n)}) \\ 
&=& \frac{1}{N}\sum_{n=1}^N (-p(X_v=k_v|x_{N_v}^{(n)})p(X_v=l_v|x_{N_v}^{(n)}))f_{k}(k_v,x_{N_v}^{(n)})f_{l}(l_v,x_{N_v}^{(n)})\;.
\end{array}
\]
 Let $W^{n,v}=(f_{j}(j_v,x_{N_v}^{(n)}), j\in J^{PS_v})$ be the $d_v\times 1$ vector of indicators. 
We introduce the notation
\begin{equation}
\eta^{n,v}_{k,l}(\t,x_{N_v}^{(n)})=\begin{cases} \frac{ \exp z_{k_v}(\theta)}{ Z^{n,v}(\t)} -(\frac{ \exp z_{k_v}(\theta)}{ Z^{n,v}(\t)})^2
, & \mbox{if } k_v=l_v\\
-\frac{ \exp z_{k_v}(\theta) \exp z_{l_v}(\theta)}{ (Z^{n,v}(\t))^2} 
, & \mbox{if } k_v\not =l_v\;.
\label{cases}
\end{cases}
\end{equation}
 Let $H^{n,v}(\t,x_{N_v}^{(n)})$ be the $d_v\times d_v$ matrix with $(k,l)$ entry $\eta^{n,v}_{k,l}(\t,x_{N_v}^{(n)})$. Then the Fisher information matrix derived from $l^{PS_v}$ is
\begin{eqnarray*}
(k^{PS_v})^{''}(\t)=\frac{1}{N}\sum_{n=1}^N H^{n,v}(\t,x_{N_v}^{(n)})\circ [W^{n,v}(W^{n,v})^t]
\end{eqnarray*}
where $\circ$ denotes the  Hadamard product of two matrices.
We make two assumptions on the behaviour of the cumulant generating function $k^{PS_v}, v\in V$ at $\theta^*$, similar to those made by Ravikumar et al. (2010) and Meng (2014):
\begin{enumerate}
	\item[(A)]  For the design matrix of the $v$-local conditional models, we assume that there exists $D_{max}>0$ such that
	$$\max_{v \in V} \lambda_{max}\Big(\frac{1}{N} \sum_{n=1}^N W^{n,v}(W^{n,v})^t\Big)\leq D_{max};$$
	\item[(B)] We assume the minimum eigenvalue of the Fisher Information matrices $(k^{PS_v})^{''}(\theta^*),\;v\in V$  is bounded, i.e., there exists $C_{min}>0$ such that
	$$C_{min}=\min_{v \in V}\lambda_{min} \frac{1}{N}\sum_{n=1}^N\big[H^{n,v}(\t^*,x_{N_v}^{(n)})\circ[ W^{n,v}(W^{n,v})^t]\big ].$$
\end{enumerate}
We are now ready to state our theorem on the asymptotic behaviour of the composite mle $\hat{\theta}$ obtained by averaging the mle of $\hat{\theta}_j^{PS_v}$ obtained from the various $v$-neighbourhood as indicated in Section 4.4.
\begin{theorem}
\label{mbound}
Assume conditions (A) and (B) hold. If the sample size N and  $|V|=p$  satisfy
\[
\frac{N}{ \log {p}} \geq \max_{v\in V}\ (\frac{10 CD_{max}d_v}{C_{min}^2})^2,
\]
where C is a positive constant such that $p^{\frac{C^2}{2}}\geq 2\sum_{v\in V}d_v$,
then the conditional composite mle $\hat{\theta}=(\hat{\theta}_j, j\in J)$  is such that
\begin{equation}
\lVert \hat{\theta}-\theta^* \rVert_F \leq \frac{5C}{C_{min}} \sqrt{\frac{\sum_{v \in V}d_v \log p}{N}}
\end{equation}
with probability greater than $1-\frac{2\sum_{v\in V}d_v}{p^{\frac{C^2}{2}}}$.
\end{theorem}
The proof is given in the Appendix. With a similar argument, we can derive the behaviour of the global mle, which we will denote by $\hat{\theta}^G$. We need to make assumptions similar to (A) and (B). We assume that
\begin{eqnarray*}
&&(A')\;\;\mbox{there exists}\;\;D_{max}>0\;\;\mbox{such that}\;\;\lambda_{max}\Big(\sum_{i\in {I}}f_i\otimes f_i\Big)\leq D_{max},\\
&&(B')\;\;0< \kappa^*=\lambda_{min}\Big[ k^{''}(\theta^*)\Big].
\end{eqnarray*}
The asymptotic behaviour of $\hat{\theta}^G$ is given in the following theorem.
\begin{theorem}
\label{global}
Assume conditions $(A')$ and $(B')$ hold. If $N$ and $p$ satisfy the condition
\[
\frac{N}{\log p} \geq (\frac{40 C|J|D_{max}}{\kappa^{*2}})^2,
\]
where C is a positive constant such that $p^{2C^2}\geq 2|J|$,
then the global mle $\hat{\theta}^G=(\hat{\theta}_j^G, j\in J)$ is such that
\begin{equation}
\lVert \hat{\theta}^G-\theta^* \rVert_F \leq \frac{5C}{\kappa^*}\sqrt{\frac{|J|\;\log{p}}{N}}
\end{equation}
with probability greater than $1-\frac{2|J|}{p^{2C^2}}$.
\end{theorem}
The proof is provided in the Supplementary file.
Comparing Theorem \ref{mbound} and \ref{global}, we see that for $\frac{N}{p}=\mathcal{O}(|J|^2)$, 
 $\lVert \hat{\theta}^G-\theta^* \rVert_F=
\mathcal{O}(\sqrt{\frac{|J|\;\log{p}}{N}})$ with high probability while for  $\frac{N}{p}=\mathcal{O}( \max_{v\in V}(d_v^2))$,
 $\lVert \hat{\theta}-\theta^* \rVert_F=\mathcal{O}(\sqrt{\frac{\sum_{v \in V}d_v \log p}{N}})$. This implies that for the composite mle, the requirement on the sample size $N$ are not as stringent as for the global mle but of course, we lose some accuracy in the approximation of $\theta^*$. The situation is, however, not bad since
$$\sqrt{\frac{\sum_{v \in V}d_v \log p}{N}}\Big/\sqrt{\frac{|J|\;\log{p}}{N}}=\sqrt{\frac{\sum_{v \in V}d_v}{|J|}}$$
which is the square root of the ratio of the sum over $v\in V$ of the number of parameters in the $v$-local conditional models and the number of parameters in the global model. If the number of neighbours for each vertex is bounded by $d$, we see that this ratio is at most equal to $\frac{2^{d+1}}{|J|}$ and usually much smaller than that. For example, in an Ising model, $|J|=p+|E|$ and $\sum_{v\in V}d_v=p+2|E|$ and therefore $\frac{\sum_{v \in V}d_v}{|J|}=1+\frac{|E|}{p+|E|}\leq 2.$

%a lattice with d neighbours to each point, $d_v\leq d+1$ and $\frac{\sum_{v \in V}d_v}{|J|}<\frac{p(d+1)}{|J|}$, which can be controlled with $d$ and grows only with $p$ as does $|J|$.
\section{Conclusion}
In this paper, we have taken a local approach to study the existence of the maximum likelihood estimate of the canonical parameter $\theta$ in a high-dimensional discrete graphical model. We have shown that we can use smaller graphical models to detect the nonexistence of this mle. We have also taken a local approach to the estimation of $\theta$ by looking at various possible versions of the composite likelihood estimate, based on local conditional or marginal likelihoods and we have shown that the two approaches yield the same estimate of $\theta$.  Through a numerical experiment, we have illustrated how we can be led to an incorrect maximum composite likelihood estimate of $\theta$ if the   global estimate does not exist. Finally, we have described the asymptotic behaviour of the maximum composite likelihood estimate of $\theta$ when both the dimension $p$ of the model and the sample size $N$ tend to infinity.
We have shown that when the number of neighbours of each $v\in V$ is bounded, the rate of convergence of the composite mle is comparable to that of the global mle.

\section{Appendix}
\subsection{Proof of  Lemma \ref{one}}
We will  use the notation $j\tl_0 j'$ to mean that $j\tl j'$ or  $j=0$, the zero cell. Let  $p^{\M_v}(i)$ denote the marginal probability of $i\in I_{\M_v}$. We know that the $\M_v$-marginal distribution of $X_{\M_v}$ is multinomial.
% but we do not know whether it is graphical or even hierarchical. However b
By the general parametrization of the multinomial model \eqref{gentheta},
%is the following: for $E\subset V$,
%$$\theta(i_E,0_{V\setminus E})=\sum_{D\subset E}(-1)^{E\setminus D}\log \frac{p^{\M_v}(i_D,i_{V\setminus D})}{p^{\M_v}(0)}.$$
for $j\in J, S(j)\subset \M_v$, since $S(j)$ is complete, 
  \be\label{thetajnv}
 \theta_j^{\M_v}=\sum_{j'\in J,\; j'\tl j}(-1)^{|S(j)|-|S(j')|}\log \frac{p^{\M_v}(j')}{p^{\M_v}(0)},
 \ee
where by abuse of notation, $j$ such that $S(j)\subset \M_v$ is considered as an element of $I_{\M_v}$.

%\subsection{Proof of  Lemma \ref{one}}
%We will  use the notation $j\tl_0 j'$ to mean that $j\tl j'$ or  $j=0$, the zero cell.
%By definition \eqref{thetaj} of the loglinear parameters and denoting $p^{\M_v}(j)$ the marginal probability of $j\in I_{\M_v}$, we have 
%  \be\label{thetajnv}
 %\theta_j^{\M_v}=\sum_{j'\in J^{\N_v},\; j'\tl j}(-1)^{|S(j)|-|S(j')|}\log \frac{p^{\M_v}(j')}{p^{\M_v}(0)}.
% \ee
%Also,
Moreover,
\begin{eqnarray*}
p^{\M_v}(j)&=&\sum_{i\in \I:\; i_{\M_v}=j}p(i)
=   \sum_{i\in \I,\; i_{\M_v}=j}\exp \{\sum_{j'\;|\;j'\tl_0 j}\theta_{j'}+\sum_{{j'\;|\;j'\tl i \atop j'\not \tl j} \atop j'_{\M_v}\tl_0 j}\theta_{j'} \}              \\
&=&\Big(\exp \sum_{j'\;|\;j'\tl_0 j}\theta_{j'}\Big)\Big(1+\sum_{i\in \I,\; i_{\M_v}=j}\exp \sum_{{j'\;|\;j'\tl i \atop j'\not \tl j} \atop j'_{\M_v}\tl_0 j}\theta_{j'} \Big)\;.
\end{eqnarray*}
Therefore 
%\begin{eqnarray*}
$\;\log p^{\M_v}(j)= \sum_{j'\;|\;j'\tl_0 j}\theta_{j'}+\log \Big(1+\sum_{i\in \I,\; i_{\M_v}=j}\exp \sum_{{j'\;|\;j'\tl i \atop j'\not \tl j} }\theta_{j'} \Big)\;,$
%\end{eqnarray*}
which we can write
\begin{eqnarray}
\label{oneway}
 \sum_{j'\;|\;j'\tl_0 j}\theta_{j'}&=&\log p^{\M_v}(j)-\log \Big(1+\sum_{i\in \I,\; i_{\M_v}=j}\exp \sum_{{k\;|\;k\tl i \atop k\not \tl j} }\theta_{k} \Big)\;.
\end{eqnarray}
Moebius inversion formula states that for $a\subseteq V$ an equality of the form
$\sum_{b\subseteq a}\Phi(b)=\Psi(a)$ is equivalent to $\Phi(a)=\sum_{b\subseteq a}(-1)^{|a\setminus b|}\Psi(b)$. Here, using a generalization of the Moebius inversion formula to the partially ordered set given by $\tl$ on $J$, we derive from  \eqref{oneway} that for $j\in J^{\M_v}\subset J$
\begin{eqnarray}
 \theta_{j}&=&\sum_{j'\;|\;j'\tl_0 j}(-1)^{|S(j)-S(j')|}\log p^{\M_v}(j')\nonumber\\
 &&-\sum_{j'\;|\;j'\tl_0 j}(-1)^{|S(j)-S(j')|}\log \Big(1+\sum_{i\in \I,\; i_{\M_v}=j'}\exp \sum_{{k\;|\;k\tl i \atop k\not \tl j'} }\theta_{k} \Big)\nonumber\\
 &=&\theta_{j}^{\M_v}-\sum_{j'\;|\;j'\tl_0 j}(-1)^{|S(j)-S(j')|}\log \Big(1+\sum_{i\in \I,\; i_{\M_v}=j'}\exp \sum_{{k\;|\;k\tl i \atop k\not \tl j'} }\theta_{k} \Big)\label{jprime}
\end{eqnarray}
which we prefer to write as \eqref{main}.
\vspace{2mm}

\subsection{Proof of Lemma \ref{two}}
Since \eqref{main} is already proved, {\it (2.)} holds. Let us  prove that {\it (1.)} holds, i.e., that when $S(j)\not \subset\B_v$, the alternating sum on the right-hand side of \eqref{main}  is equal to 0. Since $j\in J$, $S(j)$ is necessarily complete and $j'\tl j$ is obtained by removing one or more vertices from $S(j)$.

%If $S(j)\cap \B_v=\emptyset$ since any $i\in \I$ such that $i_{\M_v}=j_{\M_v}$ is such that $i_w=0, w\in \M_v,w\not\in S(j)$, the $\theta_k, k\tl i, k\not \tl j'$ in \eqref{main} are the same for all $j'\tl j$. So, the log terms in the alternating sum of logarithms in \eqref{main} are identical and cancel each other out and therefore $\theta^{\M_v}_{j_{\M_v}}=\theta_j$.

If $S(j)\cap \B_v\not =\emptyset$ but $S(j)\not \subset \B_v$, there is at least one vertex $w\in S(j)$ which is not in $\B_v$. Let $l_0$  and $l_{w}$ be the log terms in the alternating sum corresponding to $j'=0$ and $j_w'\tl j$ such that $S(j_w')=\{w\}$ respectively. Since for any  neighbours $u$ of $w$ in $\M_v$ and for any  $i\in I$ such that $i_{\M_v}=j'$, the $u$-th coordinate $i_u$ must be zero and since $w$ cannot have a neighbour outside $\M_v$, the set $\{\theta_k, k\tl i^{(1)}, k\not \tl j'\}$ in $l_0$ for $i^{(1)}$ such that $i^{(1)}_{\M_v}=0$ is the same as the set 
$\{\theta_k, k\tl i^{(2)}, k\not \tl j'\}$ in $l_w$ for $i^{(2)}$ such that $i^{(2)}_{\M_v}=j'_w$ and $i^{(2)}_{V\setminus \M_v}=i^{(1)}_{V\setminus \M_v}$. 
 The terms in $l_0$ and $l_{w}$ in \eqref{main}  are therefore exactly the same except for their sign and these two terms  cancel out. 
Similarly,  for any given $j'\tl j$ with $w\not\in S(j')$, let $j'_w\in J$ be such that $S(j'_w)=S(j)\cup \{w\}$ and $j'_w\tl j$, then,  the set $\theta_k, k\tl i^{(1)}, k\not \tl j'$ in $l_{j'}$ and the set $\theta_k, k\tl i^{(2)}, k\not \tl j'_w$ in $l_{j'_w}$ are identical where, similarly to the argument above,
$i^{(1)}$ is such that $i^{(1)}_{\M_v}=j'$ and $i^{(2)}$ is such that $i^{(1)}_{\M_v}=j'_w$ and $i^{(2)}_{V\setminus \M_v}=i^{(1)}_{V\setminus \M_v}$. 
Therefore the terms $l_{j'}$ and $l_{j'_w}$ cancel out and {\it (1.)} is proved.

 To prove that {\it (3.)} holds, following \eqref{gentheta}, we have, for $S(i)=E\subset \M_v$
 \begin{eqnarray}
 \theta_i^{\M_v}&=&\sum_{F\subset E}(-1)^{|E\setminus F|}\log p^{\M_v}(i_F,0_{\M_v\setminus F})\nonumber\\
 &=&\sum_{F\subset E}(-1)^{|E\setminus F|}\log \Big(p(i_F,0_{V\setminus F})+\sum_{L\subset V\setminus \M_v}\sum_{k_L\in I_L}p(i_{F},0_{\M_v\setminus F},k_L,0_{V\setminus (\M_v\cup L)})\Big)\nonumber\\
 &=&\sum_{F\subset E}(-1)^{|E\setminus F|}\log \Big(\exp (\sum_{j\in J, j\tl i_F}\theta_j)+\sum_{L\subset V\setminus F}\sum_{k_L\in I_L}\exp (\sum_{j\in J, j\tl i_F}\theta_j+\sum_{j\not \tl i_F, j \tl (i_{F},k_{ L})}\theta_j)\Big)\nonumber\\
 &=&\sum_{F\subset E}(-1)^{|E\setminus F|}\log \Big(\exp (\sum_{j\in J, j\tl i_F}\theta_j)\Big)\\
 &&+\sum_{F\subset E}(-1)^{|E\setminus F|}\log (1+\sum_{L\subset V\setminus F}\sum_{k_L\in I_L}\exp (\sum_{j\not \tl i_F, j \tl (i_{F},k_L)}\theta_j)\Big)\nonumber\\
 &=&\theta_i+\sum_{F\subset E}(-1)^{|E\setminus F|}\log (1+\sum_{L\subset V\setminus F}\sum_{k_L\in I_L}\exp (\sum_{j\not \tl i_F, j \tl (i_{F}, k_L)}\theta_j)\Big)\label{rhs}
 \end{eqnarray}
 Now, following an argument similar to that of {\it (1.)} above, we can show that the second component of the sum in \eqref{rhs} is equal to zero. It follows that when $\theta_i=0$ then $\theta_i^{\M_v}=0$. This completes the proof of Lemma \ref{two}.
\vspace{2mm}

\subsection{Proof of Theorem \ref{equal}}
The local relaxed marginal loglikelihood is
\begin{eqnarray*}
l^{\M_{l,v}}(\theta^{\M_{l,v}})&=&\sum_{k=1}^N\log p^{\M_{l,v}}(X_{\M_v}=i_{\M_v}^{(k)})=\sum_{i_{\M_v}\in I_{\M_v}}n(i_{\M_v})\log p^{\M_{l,v}}(i_{\M_v})\\
&=& \langle \theta^{\M_{l,v}},t^{\M_{l,v}}\rangle -Nk^{\M_{l,v}}(\theta^{\M_{l,v}})
\end{eqnarray*}
%$\square$
It is immediate to see that $\frac{\partial l^{\M_{l,v}}(\theta^{\M_{l,v}})}{\partial \theta_j}=t(j)-p^{\M_{l,v}}(j_{S(j)})$ where $p^{\M_{l,v}}(j_{S(j)})$ denotes  the $j_{S(j)}$-marginal cell probability in the ${\M_{l,v}}$-marginal model. Therefore the likelihood equations $\frac{\partial l^{\M_{l,v}}(\theta^{\M_{l,v}})}{\partial \theta_j}=0,\;j\in J^{\M_{l,v}}$ yield
\begin{eqnarray}
t(j)-p^{\M_{l,v}}(j_{S(j)})=0, 
\end{eqnarray}
where $t(j)=n(j_{S(j)})$.

For the argument to follow is essentially the same for the one-hop or two-hop neighbourhood. We present it for the more general case of the two hop neighbourhood.
 The local conditional log likelihood is
%\begin{eqnarray}
\begin{align}
l^{v,2PS}(\theta^{v,2PS})&=
%\sum_{i_{\M_v}\in I_{\M_v}}n(i_{\M_v})\log p(X_v=i_v|X_{\N_v}=
\sum_{i_{\M_v}\in I_{\M_v}}n(i_{\M_v})\log \frac{p(X_v=i_v,X_{\N_v}=i_{\N_v},X_{\N_{2v}}=i_{\N_{2v}})}{p(X_{\N_{2v}}=i_{\N_{2v}})}
\nonumber\\
&=
\sum_{i_{\M_v}\in I_{\M_v}}n(i_{\M_v})\log \frac{p^{\M^{2,v}}(X_{\M_v}=i_{\M_v})}{p^{\M^{2,v}}(X_{\N_{2v}}=i_{\N_{2v}})}\nonumber\\
&=\sum_{i_{\M_v}\in I_{\M_v}}n(i_{\M_v})\log p^{\M^{2,v}}(X_{\M_v}=i_{\N_v})-\sum_{i_{\N_{2v}}\in I_{\N_{2v}}}n(i_{\N_{2v}})\log p^{\M^{2,v}}(X_{\N_{2v}}=i_{\N_{2v}})\nonumber\\
&=l^{\M_{2,v}}(\theta^{\M_{2,v}})-\sum_{i_{\N_{2v}}\in I_{\N_{2v}}}n(i_{\N_{2v}})\log \sum_{x_{v\cup \N_v}\in I_{v\cup \N_v}}p^{\M^{2,v}}(X_{v\cup \N_v}=x_{v\cup \N_v},X_{\N_{2v}}=i_{\N_{2v}})\nonumber\\
&=l^{\M_{2,v}}(\theta^{\M_{2,v}})-Q
\end{align}
where
\begin{eqnarray}
Q&=&\sum_{i_{\N_{2v}}\in I_{\N_{2v}}}n(i_{\N_{2v}})\log \sum_{x_{v\cup \N_v}\in I_{v\cup \N_v}}\exp \Big(\theta_0+\sum_{k\tl (x_{v\cup \N_v}, i_{\N_{2v}})\atop k\in J^{\M_{2,v}}}\theta_k\Big)\label{notsame2}
\end{eqnarray}
and
$\theta_0=-\log (\sum_{i_{\M_v}\in I_{\M_v}}\exp {\sum_{k\tl i_{\M_v}, k\in J^{\M_{2,v}}}\theta_k}).$
The second equality above is due to the fact that in the expression \eqref{4.3} of $\frac{p(X_v=i_v,X_{\N_v}=i_{\N_v},X_{\N_{2v}}=i_{\N_{2v}})}{p(X_{\N_{2v}}=i_{\N_{2v}})}$,  the $\theta_j$ such that $S(j)\not \in \M_v$ and  the $\theta_j$ such that $ S(j)\subset \N_{2v}$ cancel out at the numerator and denominator and it therefore does not matter, for the conditional distribution of $X_{v\cup \N_v}$ given $X_{\N_{2v}},$ what the relationship between the neighbours are. The only thing that matters is the relationship between  the  vertices in $v\cup \N_v$ and the vertices in $\M_v$ and according to Lemma \ref{two}, that remains unchanged when we change from the global model to the  $\M_{2,v}$-marginal models.

We will now differentiate the expression of $l^{v,2PS}$ in \eqref{notsame2} with respect to $\theta_j, j\in J^{\M_{2,v}}$. We first note that 
$$\frac{\partial \theta_0}{\partial \theta_j}=p^{\M^{2,v}}(j_{S(j)}).$$
If we use the notation
$${\bf 1}_{j\tl (x_{v\cup \N_v},i_{\N_{2v}})}=\left\{\begin{array}{cc}1&\mbox{if}\;j\tl (x_{v\cup \N_v},i_{\N_{2v}})\\0&\mbox{otherwise}\end{array}\right .\;,
$$
and the notation $p^{\M_{2,v}}(i_E),\;E\subset \M_v$ to denote the marginal probability of $X_E=i_E$ in the $\M_{2,v}$-marginal model,
we have
\begin{eqnarray*}
\frac{\partial Q}{\partial \theta_j}&=&\sum_{i_{\N_{2v}}\in I_{\N_{2v}}}n(i_{\N_{2v}})\frac{\sum_{x_{v\cup \N_v}\in I_{v\cup \N_v}}
p^{\M^{2,v}}(x_{v\cup \N_v},i_{\N_{2,v}})\Big({\bf 1}_{j\tl (x_{v\cup \N_v},i_{\N_{2v}})}-p^{\M^{2,v}}(j_{S(j)})\Big)}{p^{\M^{2,v}}(i_{\N_{2,v}})}.
\end{eqnarray*}
If $j\in J^{\M_{2,v}}$ is such that $S(j)\subset \N_{2v}$, then ${\bf 1}_{j\tl (x_{v\cup \N_v},i_{\N_{2v}})}={\bf 1}_{j_{\N_{2v}}\tl i_{\N_{2v}}}$ and 
\begin{eqnarray*}
\frac{\partial Q}{\partial \theta_j}&=&\sum_{i_{\N_{2v}}\in I_{\N_{2v}}}n(i_{\N_{2v}})\frac{p^{\M^{2,v}}(i_{\N_{2v}})\Big({\bf 1}_{j_{\N_{2,v}}\tl i_{\N_{2v}}}-p^{\M^{l,v}}(j_{S(j)})\Big)}{p^{\M^{2,v}}(i_{\N_{2v}})}\\&=&\sum_{i_{\N_{2v}}\in I_{\N_{2v}}}n(i_{\N_{2v}})\Big({\bf 1}_{j_{\N_{2,v}}\tl i_{\N_{2,v}}}-p^{\M^{2,v}}(j_{S(j)})\Big)\\
&=&n(j_{S(j)})-Np^{\M^{2,v}}(j_{S(j)})
\end{eqnarray*}
At the mle of the local $\M_{l,v}$ model, from standard likelihood equations (see Lauritzen, 1996, Theorem 4.11), we have
$\hat{p}^{\M^{l,v}}(j_{S(j)})=\frac{n(j_{S(j)})}{N}$ and therefore
\begin{equation}
\label{firstq}
\frac{\partial Q}{\partial \theta_j}=0, \;\;j\in J^{\M_{2,v}},\; S(j)\subset \N_{2v}.
\end{equation}
If $j\in J^{\M_{2,v}}$ is such that $S(j)\not \subset  \N_{2v}$, i.e. if $j\in J^{v,2PS}$,
\begin{eqnarray*}
\frac{\partial Q}{\partial \theta_j}&=&\sum_{i_{\N_{2v}}\in I_{\N_{2v}}}n(i_{\N_{2v}})\frac{p^{\M^{2,v}}(j_{S(j)\cap(v\cup \N_v)},i_{\N_{2v}}){\bf 1}_{j_{\N_{2v}}\tl i_{\N_{2v}}}-p^{\M^{2,v}}(j_{S(j)})p^{\M^{2,v}}(i_{\N_{2v}})
}{p^{\M^{2,v}}(i_{\N_{2v}})}\\
&=&-p^{\M^{2,v}}(j_{S(j)})\sum_{i_{\N_{2v}}\in I_{\N_{2v}}}n(i_{\N_{2v}})+
\sum_{i_{\N_{2v}}\in I_{\N_{2v}}}\frac{n(i_{\N_{2v}})}{p^{\M^{2,v}}(i_{\N_{2v}})}p^{\M^{2,v}}(j_{S(j)\cap(v\cup \N_v)},i_{\N_{2v}}){\bf 1}_{j_{\N_{2v}}\tl i_{\N_{2v}}}
\end{eqnarray*}
Since in the $\M_{2,v}$-marginal model, all the vertices in $\N_{2,v}$ are connected by construction, at the mle of the local $\M_{2,v}$ model, $\hat{p}^{\M^{2,v}}(i_{\N_{2v}})=\frac{n(i_{\N_{2v}})}{N}$ and therefore
\begin{eqnarray}
\frac{\partial Q}{\partial \theta_j}&=&-Np^{\M^{2,v}}(j_{S(j)})+N\sum_{i_{\N_{2v}}\in I_{\N_{2v}}}p^{\M^{2,v}}(j_{S(j)\cap(v\cup \N_v)},i_{\N_{2v}}){\bf 1}_{j_{\N_{2v}}\tl i_{\N_{2v}}}\nonumber\\
&=&-Np^{\M^{2,v}}(j_{S(j)})+Np^{\M^{2,v}}(j_{S(j)})=0\label{secondq}
\end{eqnarray}
It follows from \eqref{firstq} and \eqref{secondq} that the $2PS$ component of $\hat{\theta}^{\M_{2,v}}$, i.e. 
$$\hat{\theta}_j^{\M_{2,v}}, j\in J^{2,PS}$$
is the mle of the local two-hop conditional likelihood. We therefore have
$$\hat{\theta}^{v,2PS}=(\hat{\theta}^{\M_{2,v}})_{2PS}.$$

\vspace{2mm}

\subsection{Proof of Theorem \ref{mbound}}
To prove  Theorem \ref{mbound}, we need two preliminary results.
\begin{lemma}
\label{biglemma}
 Let  $\t^{v,*}=(\theta^*)^{PS_v}$ be  the true value of the parameter for the conditional model of $X_v$ given $X_{N_v}$, and let $\hat{\t}^{PS_v}$ be the value of $\theta^{PS_v}$ that maximizes $l^{PS_v}(\t)$. Then, for $t_{J^{PS_v}}$ as in \eqref{yv}, if there exists $\epsilon>0$ such that
\begin{equation}
\lVert t_{J^{PS_v}}-(k^{PS_v})^{'}(\t^{v,*}) \rVert_{\infty} \leq \epsilon \leq\frac{C_{min}^2}{10D_{max}d_v}
\end{equation}
then
\begin{equation}
\lVert \hat{\theta}^{PS_v}-\theta^{v,*} \rVert_F \leq \frac{5\sqrt{d_v} \epsilon}{C_{min}}
\end{equation}
\end{lemma}
\noindent {\bf Proof.}
To simplify our notation in this proof, we drop any  subscripts and superscripts containing $v$ or $PS$, except when it is necessary to keep them to make the argument clear.

Let $Q(\Delta)=l(\theta^*)-l(\theta^* +\Delta)$. Clearly  $Q(0)=0$ and $Q(\hat{\Delta}) \leq Q(0)=0$, where $\hat{\Delta}=\hat{\theta}-\theta^*$. Let $||\Delta||_F=\sqrt{\sum_{j\in J^{PS_v}}\Delta_j^2}$ denote the Frobenius norm of $\Delta.$ 
Define $C(\delta)=\{\Delta | \quad \lVert \Delta \rVert_F=\delta \}$. Since $Q(\Delta)$ is a convex function of $\Delta$, if we can prove 
\begin{equation}
\label{positive}
\inf_{\Delta \in C(\delta)} Q(\Delta) > 0,
\end{equation}
then, by convexity of $Q$, it will follow that $\hat{\Delta}$ must lie within the sphere defined by $C(\delta)$, i.e. $ \rVert\hat{\Delta} \rVert_F\leq \delta.$
%\[
%\lVert \hat{\Delta} \rVert_2\leq \delta.
%\]
We are now going to prove that there exists $\delta>0$ such that  on $C(\delta)$, $Q(\Delta)>0$.
%within which $Q(\delta)>0$.
For $\Delta \in C(\delta)$, we have
\begin{equation}
\label{Q1}
\begin{array}{lcl}
Q(\Delta)&=&l(\theta^*)-l(\theta^*+\Delta) = \t^{*t}t - k(\t^{*})-((\t^{*}+\Delta)^{t}t - k(\t^{*}+\Delta))\\
&=& k(\theta^*+\Delta)-k(\theta^*)-\Delta^tt=\Delta^tk^{'}(\theta^*)+ \frac{1}{2}\Delta^{t}k^{''}(\theta^*+ \alpha \Delta) \Delta-\Delta^tt, \quad \alpha \in [0,1]\\
&=&\underbrace{\Delta^t[k^{'}(\t^*)-t]}_{Q_1} + \underbrace{\frac{1}{2}\Delta^{t}k^{''}(\theta^*+ \alpha \Delta) \Delta}_{Q_2}
\end{array}
\end{equation}
%Take the Taylor expansion of $k(\theta^*+\Delta)$ around $\theta^*$, we can get
%\[
%k(\theta^*+\Delta)-k(\theta^*) =\Delta^tk^{'}(\theta^*)+ \frac{1}{2}\Delta^{t}k^{''}(\theta^*+ \alpha \Delta) \Delta, \quad \alpha \in [0,1]
%\]
By Cauchy-Schwartz inequality, we have the following bound for $Q_1$.
\begin{equation}
\label{Q11}
\begin{array}{lcl}
|Q_1|=|\Delta^t[k^{'}(\t^*)-t]| &\leq& \lVert k^{'}(\theta^*) - t\rVert_{\infty} || \Delta ||_1 
\leq \epsilon \sqrt{d}\lVert \Delta \rVert_F=\epsilon \sqrt{d}\delta
\end{array}
\end{equation}

\noindent For $Q_2$, we have
\begin{equation}
\label{Q2}
Q_2 \geq \frac{1}{2} \lVert \Delta \rVert_F^2 \min_{\alpha \in [0,1]}\lambda_{min}k^{''}(\theta^*+\alpha \Delta)=\frac{1}{2} \delta^2 \min_{\alpha \in [0,1]}\lambda_{min}k^{''}(\theta^*+\alpha \Delta) 
\end{equation}

We now want to bound the term $q= \min_{\alpha \in [0,1]}\lambda_{min} [k^{''}(\theta^*+\alpha \Delta)]$ from below. 
Following \eqref{ztheta}, we can write $z_{y_v}(\theta+\alpha\Delta)=\sum_{j \in J;v \in S(j)} (\t_j+\alpha\Delta_j) f_j(y_v,x_{N_v}^{(n)})$, then we can rewrite the entries of $H$ in \eqref{cases} as 
\[
\eta^{n,v}_{k,l}(\t^*+\alpha \Delta,x_{N_v}^{(n)})=\begin{cases} \frac{ \exp z_{k_v}(\theta^*+\alpha\Delta)}{ 1+ \sum_{y_v \in I_v \setminus \{0\}} \exp z_{y_v}(\theta^*+\alpha\Delta)} -(\frac{ \exp z_{k_v}(\theta^*+\alpha\Delta)}{ 1+ \sum_{y_v \in I_v \setminus \{0\}} \exp z_{k_v}(\theta^*+\alpha\Delta)} )^2, & \mbox{if } k_v=l_v\\-\frac{ \exp z_{k_v}(\theta^*+\alpha\Delta)\exp z_{l_v}(\theta^*+\alpha\Delta)}{ (1+ \sum_{y_v \in I_v \setminus \{0\}} \exp z_{y_v}(\theta^*+\alpha\Delta) )^2}, & \mbox{if } k_v\not =l_v
\end{cases}
\]
then 
\[
\frac{\partial\eta^{n,v}_{k,l}(\t^*+\alpha \Delta,x_{N_v}^{(n)})}{\partial \alpha}=\sum_{y_v \in I_v \setminus \{0\}} (\eta^{n,v}_{k,l})^{'}_{y_v}(\t^*+\alpha \Delta,x_{N_v}^{(n)}) \frac{\partial z_{y_v}}{\partial \alpha},
\]
where $(\eta^{n,v}_{k,l})^{'}_{y_v}(\t^*+\alpha \Delta,x_{N_v}^{(n)})= \frac{\partial\eta^{n,v}_{k,l}(\t^*+\alpha \Delta,x_{N_v}^{(n)})}{\partial z_{y_v}}$. It is easy to see that these derivatives can all be expressed in terms of probabilities of the type \eqref{p} and that they are always less than $1$ in absolute value.
%{\bf Aside} Let $p_k= \frac{ \exp z_{k_v}}{ 1+ \sum_{y_v \in \mathcal{X}_v \setminus \{0\}} \exp z_{y_v}} -(\frac{ \exp z_{k_v}}{ 1+ \sum_{y_v \in \mathcal{X}_v \setminus \{0\}} \exp z_{k_v}} )^2$ and similarly for $p_l$ and $p_y$. Then
%\begin{eqnarray}
%\eta^{'}_{k_v}(\t^*+\alpha \Delta,x_{N_v}^{(n)})&=& \frac{\partial\eta(\t^*+\alpha \Delta,x_{N_v}^{(n)})}{\partial z_{k_v}}=p_k-p_k^2-2p_k(p_k-p_k^2)=(p_k-p_k^2)(1-2p_k);\;\mbox{if}\; k_v=l_v\\
%\eta^{'}_{k_v}(\t^*+\alpha \Delta,x_{N_v}^{(n)})&=&-p_kp_l(1-2p_k);\;\mbox{if}\; k_v\not=l_v\\
%\eta^{'}_{y_v}(\t^*+\alpha \Delta,x_{N_v}^{(n)})&=&-p_kp_y(1-2p_k);\;\mbox{if}\; k_v=l_v\\
%\eta^{'}_{y_v}(\t^*+\alpha \Delta,x_{N_v}^{(n)})&=&2p_kp_lp_y;\;\mbox{if}\; k_v\not=l_v
%\end{eqnarray}
%{\bf End of Aside}
%Since the functions $(p_k,p_l)\mapsto p_k-p_k^2$ and  $(p_k,p_l)\mapsto -p_kp_l$ are always less than 1 in absolute value for $(p_k,p_l)\in [0,1]^2$, it follows immediately that $|\eta^{'}_{y_v}(\t^*+\alpha \Delta,x_{N_v}^{(n)})| \leq 1, \forall y_v \in \mathcal{X}_v \setminus \{0\}$.
 Therefore, since  $\frac{\partial z_{y_v}(\theta+\alpha \Delta)}{\partial \alpha}= \sum_{j \in J;v \in S(j)} \Delta_j f_{j}(y_v,x^n_{N_v})$
\begin{equation}
\begin{array}{lcl}
|\frac{\partial\eta^{n,v}_{k,l}(\t^*+\alpha \Delta,x_{N_v}^{(n)})}{\partial \alpha}| &\leq& \sum_{y_v \in I_v \setminus \{0\}}  \frac{\partial z_{y_v}}{\partial \alpha} = \sum_{y_v \in I_v \setminus \{0\}}  \sum_{j \in J; v \in S(j)}\Delta_j f_{j}(y_v,x^n_{N_v})\\
&=& \sum_{j \in J; v \in S(j)} \Delta_j \sum_{y_v \in I_v \setminus \{0\}}  f_{j}(y_v,x^n_{N_v}) =\<\Delta, W^{n} \>\;.
\end{array}
\end{equation}

The Taylor series expansion of $\eta^{n,v}_{k,l}(\t^*+\alpha \Delta,x_{N_v}^{(n)})$ yields
\[
\eta^{n,v}_{k,l}(\t^*+\alpha \Delta,x_{N_v}^{(n)})=\eta^{n,v}_{k,l}(\t^*,x_{N_v}^{(n)})+\alpha \frac{\partial\eta^{n,v}_{k,l}(\t^*+\alpha' \Delta,x_{N_v}^{(n)})}{\partial \alpha}, \ \alpha^{'} \in [0,\alpha]\;.
\]
Let $K(\t^*+\alpha^{'} \Delta,x_{N_v}^{(n)})$  denote the $d_v\times d_v$ matrix with entry $\frac{\partial\eta^{n,v}_{k,l}(\t^*+\alpha \Delta,x_{N_v}^{(n)})}{\partial \alpha}$.
Coming back to \eqref{Q2}, we have
\[\begin{array}{lcl}
k^{''}(\theta^*+\alpha \Delta)&=&\frac{1}{N}\sum_{n=1}^N\big[H(\t^*+\alpha \Delta,x_{N_v}^{(n)})\circ[ W^n(W^n)^t]\big ] \\
&=& \frac{1}{N}\sum_{n=1}^NH(\t^*,x_{N_v}^{(n)})\circ[ W^n(W^n)^t]+\alpha \frac{1}{N}\sum_{n=1}^NK(\t^*+\alpha^{'} \Delta,x_{N_v}^{(n)})\circ[ W^n(W^n)^t]\;.\end{array}\]
We write $||X||_2=\lambda_{max}(X)$ for the operator norm of a matrix $X$. We recall that the Hadamard product of two positive semidefinite matrices is positive semidefinite and therefore 
$$\lambda_{min}(K(\t^*+\alpha^{'} \Delta,x_{N_v}^{(n)})\circ[ W^n(W^n)^t])\geq -\lambda_{max}(K(\t^*+\alpha^{'} \Delta,x_{N_v}^{(n)})\circ[ W^n(W^n)^t]).$$ Then since $|\alpha|<1$, we have
\begin{equation}
\begin{array}{lcl}
q&=&\min_{\alpha \in [0,1]} \lambda_{min}[\frac{1}{N}\sum_{n=1}^NH(\t^*+\alpha \Delta,x_{N_v}^{(n)})W^n(W^n)^t] \\
&\geq& \lambda_{min}(\frac{1}{N}\sum_{n=1}^N\big[H(\t^*,x_{N_v}^{(n)})\circ(W^n(W^n)^t)\big ])\\
&&\hspace{2cm}-\max_{\alpha \in [0,1]} ||\alpha \frac{1}{N}[\sum_{n=1}^NK(\t^*+\alpha \Delta,x_{N_v}^{(n)})\circ(W^n(W^n)^t)]||_2 \\
&\geq& C_{min}- \max_{\alpha \in [0,1]} \lVert \underbrace{\frac{1}{N}\sum_{n=1}^N\Delta^tW^n (W^n(W^n)^t)}_{A}\rVert_2\\
&\geq& C_{min}- \max_{\alpha \in [0,1]} ||A||_2\;,\label{q}
\end{array}
\end{equation}
where the last but one inequality is due to our Assumption (B). We now need to bound the spectral norm of $A=\frac{1}{N}\sum_{n=1}^N\Delta^tW^n (W^n(W^n)^t).$
For any $\alpha \in [0,1]$ and $y \in R^{d_v}$ with $||y||_F=1$, we have
%\[
%\begin{array}{lcl}
\begin{eqnarray}
\<y,Ay\>&=&\frac{1}{N}\sum_{n=1}^N(\Delta^tW^n)(y^tW^n)^2\leq \frac{1}{N}\sum_{n=1}^N|\Delta^tW^n|(y^tW^n)^2,\nonumber\\
|\Delta^tW^n| &\leqslant& \sqrt{d}||\Delta||_F=\sqrt{d}\delta\;.\label{CS}
%\end{array}
%\]
\end{eqnarray}
and, by definition of the operator norm and from Assumption (B),  
\begin{equation}\label{AB}
\frac{1}{N}\sum_{n=1}^N(y^tW^n)^2 \leq ||\frac{1}{N} \sum_{n=1}^N W^n(W^n)^t||_2<D_{max}\;.
\end{equation}
From \eqref{q}, \eqref{CS} and \eqref{AB}, we obtain $\max_{\alpha \in [0,1]} ||A||_2\leq D_{max}\sqrt{d}\delta$ and therefore
\[
q\geq C_{min}- D_{max}\sqrt{d}\delta\;.
\]
Substituting this into \eqref{Q2}, we get 
\begin{equation}
\label{Q22}
 Q_2 \geq \frac{1}{2}\delta^2 (C_{min}- D_{max}\sqrt{d}\delta).
\end{equation}
From the two inequalities \eqref{Q11} and \eqref{Q22}, it follows that
\begin{equation}
\label{last}
Q(\Delta) \geq Q_2-|Q_1| \geq \frac{1}{2}\delta^2 (C_{min}- D_{max}\sqrt{d}\delta) - \epsilon \sqrt{d}\delta.
\end{equation}
To simplify the problem, we can choose $\delta$ such that $C_{min}- D_{max}\sqrt{d}\delta  \geq \frac{C_{min}}{2}$, that is, $\delta \leq \frac{C_{min}}{2 D_{max}\sqrt{d}}$.
Then inequality \eqref{last} becomes 
\[
Q(\Delta) \geq \frac{C_{min} \delta^2}{4}- \epsilon \sqrt{d}\delta
\]
and
$Q(\Delta)$ is positive if we let $\delta= \frac{5\sqrt{d }\epsilon}{C_{min}}$. Moreover  $\delta \leq \frac{C_{min}}{2 D_{max}\sqrt{d}}$ yields the following bound of $\epsilon$:
\[
\epsilon \leq \frac{C_{min}^2}{10D_{max}d}.
\]
We have therefore shown that \eqref{positive} holds for $\delta= \frac{5\sqrt{d }\epsilon}{C_{min}}$ and the theorem is proved.
%So, we have found a proper $\delta>0$ such that $Q(\Delta) \geq 0$, so we have
%\[
%\lVert \hat{\theta}^{ps}-\theta^* \rVert_F \leq \delta = \frac{5\sqrt{d }\epsilon}{C_{min}},
%\]
%when \[
%\lVert Y-k^{'}(\t^*) \rVert_{\infty} \leq \epsilon \leq \frac{C_{min}^2}{10D_{max}d}.
%\]
$\square$ 
In the next lemma, we will make use of Hoeffding inequality (see Hoeffding (1963), Theorem 2) which states the following. If If $X_1,X_2,\dotsm, X_n$ are independent and $a_i \leq X_i \leq b_i (i=1,2,\dotsm ,n)$, then for $\epsilon >0$
\[
p(|\bar{X}-\mu | \geq \epsilon) \leq 2\exp \big ( \frac{-2n^2\epsilon^2}{\sum_{i=1}^n (b_i-a_i)^2} \big )\;.
\]
\begin{lemma}
\label{Hoeffding}
Let $t_{J^{PS_v}}, k^{PS_v}$ and $d_v$ be as defined above. For any  $\epsilon>0$, we have
\begin{equation}
p(\{\lVert t_{J^{PS_v}}-(k^{PS_v})^{'}(\t^{v,*}) \rVert_{\infty} \geq \epsilon\}) \leq 2d_v\exp (-\frac{N\epsilon^2}{2})
\end{equation}
and we also have
\begin{equation}
p(\{\max_{v \in V}\lVert y^v-(k^{PS_v})^{'}(\t^*) \rVert_{\infty} \geq \epsilon\}) \leq 2\sum_{v \in V} d_v\exp (-\frac{N\epsilon^2}{2})
\end{equation}
\end{lemma}
\noindent {\bf Proof.}
In this proof, as in the previous proof, we drop the superscripts $v,PS$ except when necessary. For $j\in J^{PS_v}$, we clearly have
\[
E_{\t^*}\Big(\frac{\partial l(\theta)}{\partial \theta_j}\Big)=E_{\t^*}\Big(t_j-\frac{\partial k(\theta)}{\partial \theta_j}\Big)
=E_{\t^*}\Big(\frac{1}{N}\sum_{n=1}^Nf_{j}(x_v^{(n)},x_{N_v}^{(n)})-p(x_v=j_v|x^n_{N_v})f_{j}(x_v=j_v,x_{N_v}^{(n)})\Big)=0
\]
and  $|f_{j}(x_v^{(n)},x_{N_v}^{(n)})-p(x_v=j_v|x^{(n)}_{N_v})f_{j}(x_v=j_v,x_{N_v}^{(n)})|\leq 1.$ Moreover, by Hoeffding inequality,
\[
p(|t_j-k_j^{'}(\t^*)|\geq \epsilon) \leq 2\exp -\frac{2N^2\epsilon^2}{2^2N}=
2\exp (-\frac{N\epsilon^2}{2})
\]
Applying a union bound, we get, for each $v\in V$,
\[
p(\lVert t_{J^{PS_v}}-k^{'}(\t^*)\rVert_{\infty} \geq \epsilon)\leq \sum_{j=1}^{d_v} p(|t_j-k_j^{'}(\t^*)|\geq \epsilon) \leq 2d_v\exp (-\frac{N\epsilon^2}{2})
\]
from which it follows that
\[
p\{\max_{v \in V}\lVert t_{J^{PS_v}}-(k^{PS_v})^{'}(\t^*) \rVert_{\infty} \geq \epsilon\} \leq \sum_{v \in V}p(\lVert t_{J^{PS_v}}-(k^{PS_v})^{'}(\t^*)\rVert_{\infty} \geq \epsilon) \leq 2\sum_{v \in V} d_v\exp (-\frac{N\epsilon^2}{2})
\]
$\square$
\vspace{2mm}

{\bf Proof of Theorem \ref{mbound}}
\noindent {\bf Proof.}
Let $\epsilon = C\sqrt{\frac{\log {p}}{N}}$, where $C$ is a constant that we will choose later in this proof. From Lemma \ref{Hoeffding}, we have
\[
p(\max_{v \in V}\lVert t_{J^{PS_v}}-(k^{PS_v})'(\t^*)\rVert_{\infty} \geq C\sqrt{\frac{\log {p}}{N}}) \leq 2\sum_{v \in V}{d_v}\exp (-\frac{N(C\sqrt{\frac{\log {p}}{N}})^2}{2})=\frac{2\sum_{v \in V}{d_v}}{p^{\frac{C^2}{2}}}
\]

From Lemma \ref{biglemma}, for $\epsilon = C\sqrt{\frac{\log {p}}{N}} \leq \frac{C_{min}^2}{10D_{max}d_v}$, i.e. for $N \geq (\frac{10 CD_{max}d_v}{C_{min}^2})^2 \log {p}$, we have
$$\lVert t_{J^{PS_v}}-(k^{PS_v})'(\t^*) \rVert_{\infty} \leq \epsilon \leq\frac{C_{min}^2}{10D_{max}d_v} \Rightarrow \lVert \hat{\theta}^{PS_v}-\theta^{v,*} \rVert_F \leq \frac{5\sqrt{d_v} \epsilon}{C_{min}}.$$

The global composite mle $\hat{\theta}$ obtained by the local averaging of the $\hat{\theta}^{PS_v}$ from each conditional model can then be bounded as follows:
\[
\begin{array}{lcl}
\lVert \hat{\theta}-\theta^* \rVert_F &\leq& \big( \sum_{v \in V} \lVert \hat{\theta}^{PS_v}-\theta^{v,*} \rVert_F^2 \big )^{\frac{1}{2}} \\
&\leq& (\sum_{v \in V} (\frac{5\sqrt{d_v}C\sqrt{\frac{\log {p}}{N}}}{C_{min}})^2)^{\frac{1}{2}}\\
&=& \frac{5C}{C_{min}} \sqrt{\frac{\sum_{v \in V}d_v \log p}{N}}
\end{array}
\]
Therefore under the condition $N \geq \max_{v\in V}\ (\frac{10 CD_{max}d_v}{C_{min}^2})^2 \log {p}$, the following holds
\[
\begin{array}{lcl}
&&\max_{v \in V}\lVert t_{J^{PS_v}}-(k^{PS_v})^{'}(\t^*)\rVert_{\infty} \geq C\sqrt{\frac{\log {p}}{N}}) \Rightarrow \lVert t_{J^{PS_v}}-(k^{PS_v})^{'}(\t^*)\rVert_{\infty} \geq C\sqrt{\frac{\log {p}}{N}})\\
&\Rightarrow&\lVert \hat{\theta}^{PS_v}-\theta^{v,*} \rVert_F \leq \frac{5\sqrt{d_v} \epsilon}{C_{min}} \Rightarrow \lVert \hat{\theta}-\theta^* \rVert_F \leq \frac{5C}{C_{min}} \sqrt{\frac{\sum_{v \in V}d_v \log p}{N}},
\end{array}
\]
with
\[
p(\lVert \hat{\theta}-\theta^* \rVert_F \leq \frac{5C}{C_{min}} \sqrt{\frac{\sum_{v \in V}d_v \log p}{N}}) \geq p(\max_{v \in V}\lVert t_{J^{PS_v}}-k_v^{'}(\t^*)\rVert_{\infty} \leq C\sqrt{\frac{\log {p}}{N}}) \geq 1- \frac{2\sum_{v \in V}d_v}{p^{\frac{C^2}{2}}}
.\]

The theorem would make no sense if probability of the convergence rate was negative and thus $C$ must satisfy
$$1-\frac{2\sum_{v \in V}d_v}{p^{\frac{C^2}{2}}}> 0 \Rightarrow C \geq \sqrt{2\frac{\log 2\sum_{v \in V}d_v}{\log p}}\;.$$
$\square$

\end{document}